\documentclass[11pt]{article}

\usepackage[T1]{fontenc}
\usepackage{lmodern} 
\usepackage[scaled=0.95]{helvet} 

\usepackage[margin=1in]{geometry}
\usepackage{titlesec}
\usepackage{graphicx}
\usepackage{parskip}
\usepackage[dvipsnames]{xcolor}
\usepackage{tcolorbox}
\usepackage[numbers,square]{natbib}  
\usepackage{amsmath, amssymb}
\usepackage{fancyhdr}
\usepackage{microtype}
\usepackage{sectsty}
\usepackage{enumitem}
\usepackage{hyperref}  

\usepackage{url}
\usepackage{booktabs}
\usepackage{amsfonts}
\usepackage{nicefrac}
\usepackage{float}
\usepackage{algorithm}
\usepackage{algpseudocode}
\usepackage{inconsolata}
\usepackage[labelfont=bf]{caption}
\usepackage{multirow}
\usepackage{refcount}
\usepackage{footmisc}

\definecolor{metaBlue}{RGB}{24,119,242}
\definecolor{abstractBoxBG}{RGB}{245,245,250}

\fancypagestyle{firstpagestyle}{
  \fancyhf{}
  \rhead{\textsf{\small Stanford NeuroAI Lab}}
  \cfoot{\thepage}
}
\sectionfont{\sffamily\large\bfseries\color{black}}
\subsectionfont{\sffamily\normalsize\bfseries\color{black}}

\usepackage{xspace}
\newcommand{\lras}{\textbf{\texttt{LRAS}}\xspace}
\newcommand{\lrasd}{\textbf{\texttt{LRAS-3D}}\xspace}

\newcommand{\lrasseg}{$\textbf{\texttt{SpelkeNet}}$\xspace}
\newcommand{\spelkeentity}{$\textbf{\texttt{SpelkeBench}}$\xspace}

\begin{document}

\vspace*{-2em}
\begin{tcolorbox}[
  colback=abstractBoxBG,
  colframe=abstractBoxBG,  
  boxrule=0pt,
  arc=4pt,
  left=12pt, right=12pt, top=10pt, bottom=12pt,
  width=\textwidth,
  enlarge left by=0mm,
  enlarge right by=0mm
]

{\sffamily\LARGE\bfseries Discovering and using Spelke segments \\[0.5em]}
{\sffamily\textbf{Rahul Venkatesh}$^{*,1,\dagger}$, \textbf{Klemen Kotar}$^{*,1}$, \textbf{Lilian Naing Chen}$^{*,1}$, \\
\textbf{Seungwoo Kim}$^{1}$,
\textbf{Luca Thomas Wheeler}$^{1}$, \textbf{Jared Watrous}$^{1}$, 
\textbf{Ashley Xu}$^{1}$,
\textbf{Gia Ancone}$^{1}$,\\
\textbf{Wanhee Lee}$^{1}$,
\textbf{Honglin Chen}$^{2}$,
\textbf{Daniel Bear}$^{3}$, \textbf{Stefan Stojanov}$^{1}$, \textbf{Daniel Yamins}$^{1, \dagger}$} \\[0.5em]
\small
$^1$Stanford University, 
$^2$OpenAI,
$^3$Noetik Inc. \\[1em]
\textbf{Abstract.} Segments in computer vision are often defined by semantic considerations and are highly dependent on category-specific conventions. In contrast, developmental psychology suggests that humans perceive the world in terms of Spelke objects---groupings of physical things that reliably move together when acted on by physical forces. Spelke objects thus operate on category-agnostic causal motion relationships which potentially better support tasks like manipulation and planning. In this paper, we first benchmark the Spelke object concept, introducing the \spelkeentity dataset that contains a wide variety of well-defined Spelke segments in natural images. Next, to extract Spelke segments from images algorithmically, we build \lrasseg, a class of visual world models trained to predict distributions over future motions.  \lrasseg supports estimation of two key concepts for Spelke object discovery: (1) the motion affordance map, identifying regions likely to move under a poke, and (2) the expected-displacement map, capturing how the rest of the scene will move. These concepts are used for ``statistical counterfactual probing'', where diverse ``virtual pokes'' are applied on regions of high motion-affordance, and the resultant expected displacement maps are used to define Spelke segments as statistical aggregates of correlated motion statistics. We find that \lrasseg outperforms supervised baselines like SegmentAnything (SAM) on \spelkeentity. Finally, we show that the Spelke concept is practically useful for downstream applications, yielding superior performance on the 3DEditBench benchmark for physical object manipulation when used in a variety of off-the-shelf object manipulation models. Project page:  \url{https://neuroailab.github.io/spelke_net}.
\\

\textsuperscript{$\dagger$}Corresponding authors: rahulvenkk@gmail.com, dyamins@gmail.com.

\end{tcolorbox}

\renewcommand{\thefootnote}{}
\footnotetext{*Equal contribution. Author order randomly decided.}
\renewcommand{\thefootnote}{\arabic{footnote}} 

\vspace{2em}

\vspace{-2.5em}
\begin{figure}[!ht]  
  \centering
  \includegraphics[width=\linewidth]{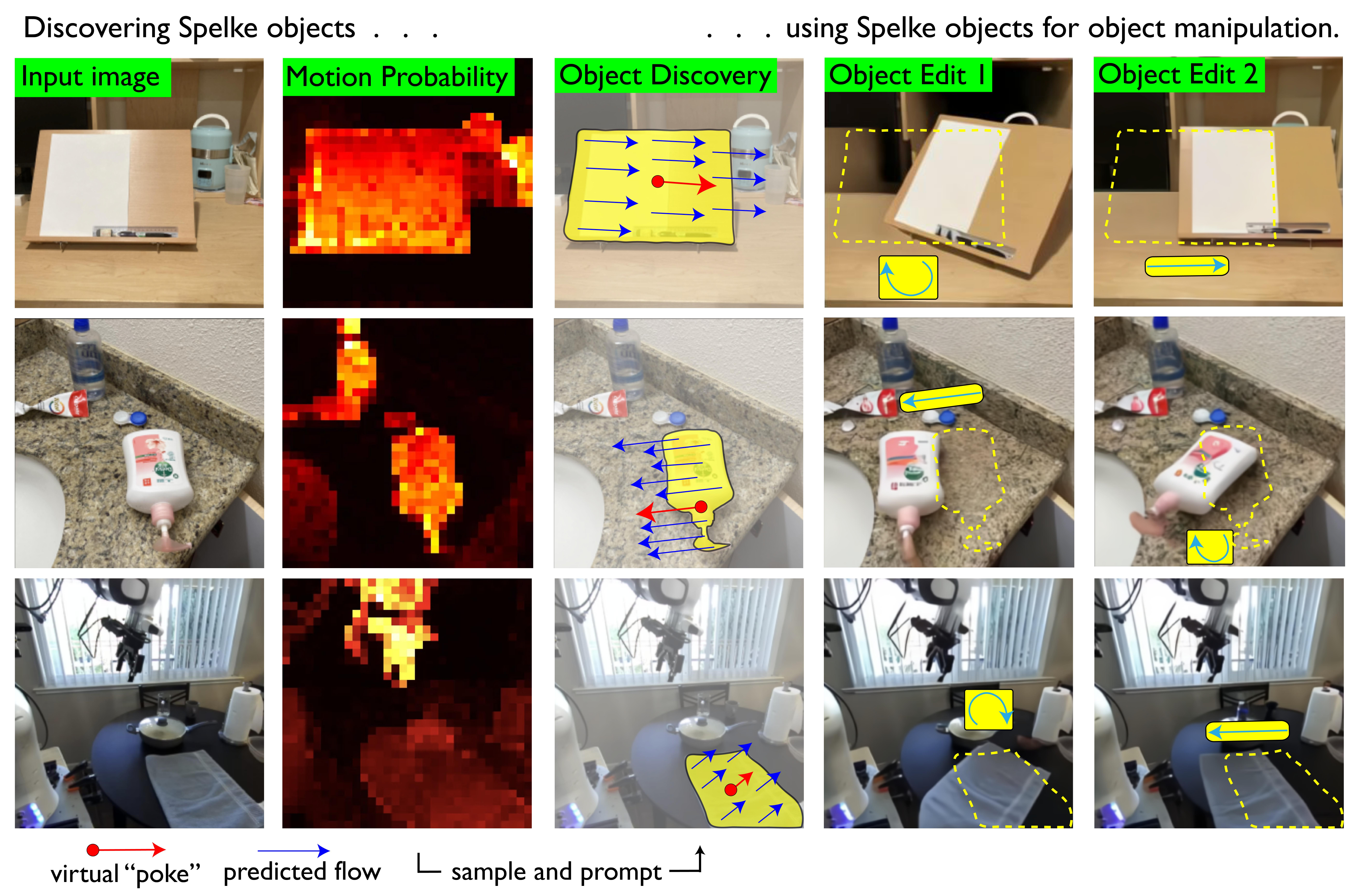}
  \vspace{-0.5em}
  \caption{
  \textbf{Overview of \lrasseg's capabilities.} Our model first predicts a probability of motion map, indicating regions likely to undergo movement independent of camera motion---i.e. candidate movable objects. We sample a point from this map and apply a virtual poke. Conditioned on this intervention, our model completes the flow field. From this, we extract a grouping of pixels, or a ``segment'' corresponding to an entity that would move as a cohesive whole under the application of external forces (i.e. a Spelke object). On the right, we illustrate how these discovered segments can be used in a physical object editing pipeline to precisely define the object we desire to manipulate---ensuring that edits are applied to groups of pixels that would move together in the real world as opposed to segments defined based on appearance or semantics. We show in this paper that the Spelke segments we discover enable more physically plausible object manipulation.}

  \label{fig:teaser}
  \vspace{-1.0em}
\end{figure}

\section{Introduction}

As the work of developmental psychologist Elizabeth Spelke and others has shown, children in their early months of life already possess notions of objecthood based on motion, segmenting the visual world into bounded units that move and interact as cohesive wholes~\cite{spelke1990principles}.  These early capabilities enable infants to track and predict how objects behave under physical forces, laying the cognitive groundwork for intuitive physical reasoning~\cite{johnson2004development,feldman1999role,brox2010object}.  

\begin{figure}[b!]
  \centering
  \includegraphics[width=\textwidth]{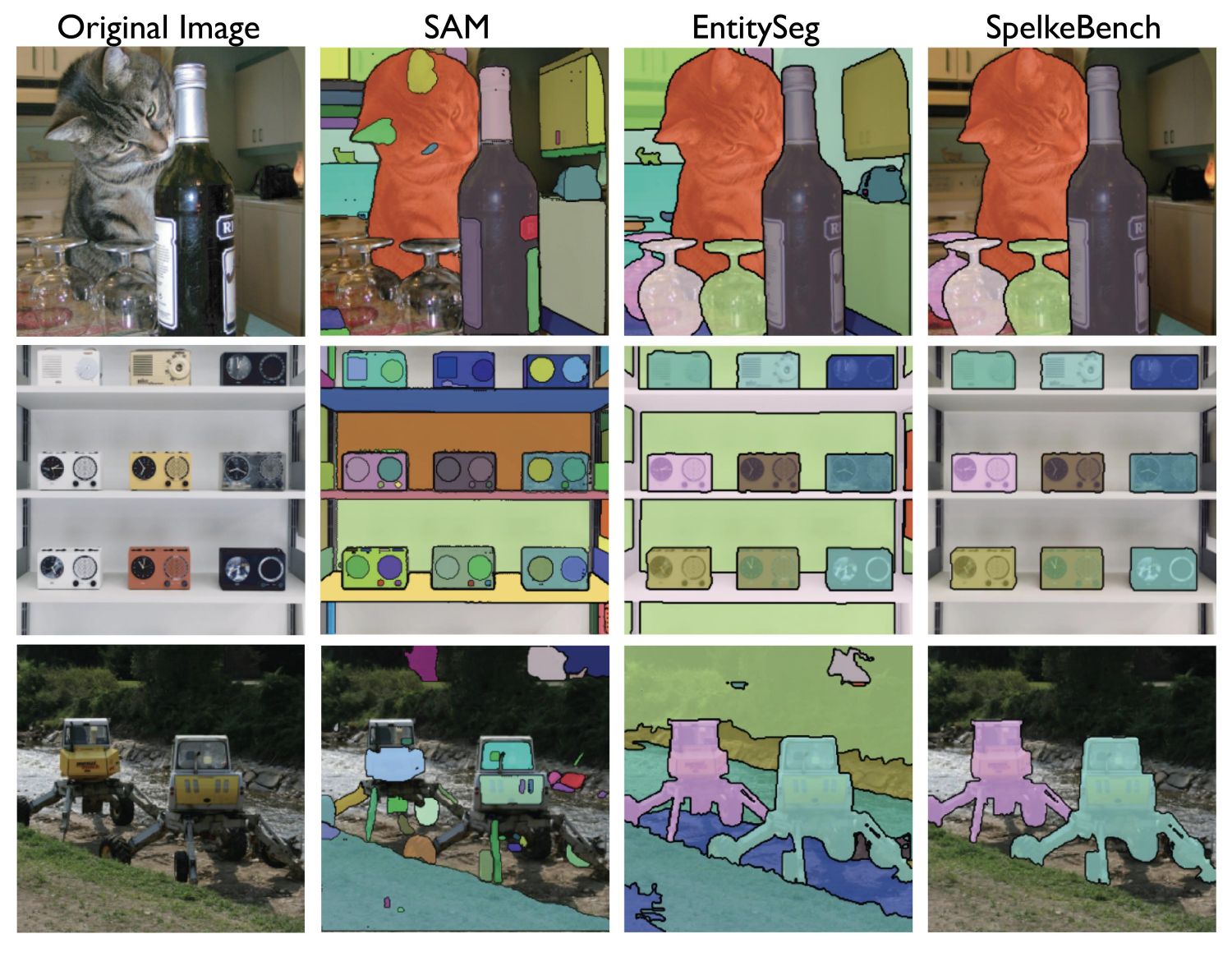}
  \caption{\textbf{Benchmarking Spelke Segmentation: comparing Spelke segments with conventional segmentation definitions.}
SAM~\cite{ravi2024sam2} produces fine-grained segments but often includes regions that do not typically move independently when forces are applied—such as logos on bottles, shadows, or sub-parts of objects like camera lenses—reflecting its focus on visual distinctiveness over physical structure. Entity segmentation~\cite{Qi_2023_EntitySeg} more closely approximates the Spelke notion of segmentation but still includes non-movable elements like walls, streets, and fixed shelves. Our \spelkeentity benchmark is constructed by manually filtering out such segments (as described in Section~\ref{sec:spelke_benchmark}), retaining only those that correspond to physically grounded entities defined by correlated motion in response to applied forces.}
  \label{fig:ours_vs_SAM}
\end{figure}

While the idea of such motion-based ``Spelke objects'' is cognitively natural, typical ontologies of segmentation in the computer vision literature diverge substantially from the Spelke concept. Standard segmentation datasets like COCO~\cite{lin2015microsoftcococommonobjects} and ADE20K~\cite{zhou2018semanticunderstandingscenesade20k} define segments based on semantic or instance-level labels such as car, tree, or amorphous categories like ``stuff''~\cite{Qi_2023_EntitySeg}. Although these are useful for recognition tasks, the resulting masks often do not reflect how objects move or interact in the real world. This highlights a core challenge in segmentation: conventional definitions may not reflect the causal structure required for physical reasoning. In contrast, Spelke segments provide a category-agnostic alternative by grouping regions based on their responses to forces, resulting in a more functional notion of segmentation, which can serve as a useful foundation for vision systems in robotics tasks like manipulation and planning (see Figure~\ref{fig:ours_vs_SAM}).



In this work, we introduce a self-supervised procedure for extracting Spelke segments from real-world static images, and show how these Spelke segments can be effectively used as the basis for physically grounded object manipulation tasks.  

We first benchmark the Spelke segment concept by introducing \spelkeentity: an evaluation dataset designed to assess whether segmentation algorithms can identify Spelke segments. This benchmark allows us to systematically measure how well a model's output align with the Spelke concept. While collecting such a dataset on a scale suitable for evaluation is feasible, collecting it on a scale suitable for supervised training is prohibitively expensive. Capturing annotations that respect Spelke's core principles---cohesion, continuity, solidity, and contact---requires nuanced human judgment and possibly some form of physical interaction, making it impractical to scale through conventional labeling pipelines.


To sidestep this bottleneck, we turn to self-supervised visual world models. Specifically, we build on the recently proposed Local Random Access Sequence Modeling (\lras)~\cite{lee20253d}, a probabilistic framework that models sequences of locally quantized tokens. Here, we introduce \lrasseg as a specific instance of \lras trained to predict a distribution of plausible future flow fields given an input image. By virtue of being trained on large-scale internet videos, \lrasseg acquires an implicit understanding of ``what moves together'' in natural scenes without ever being given explicit segmentation labels. 

The autoregressive structure and locality properties of the \lras framework are a natural architectural foundation for \lrasseg: Spelke segments can be discovered by applying localized virtual pokes simply by appending optical flow tokens to the input sequence and having the model complete the flow field to discover motion correlation patterns. In contrast, diffusion-based models~\cite{rombach2022high} rely on dense, global conditioning, making such targeted interventions hard.



To \textit{discover} Spelke segments, we introduce a ``statistical counterfactual probing'' procedure on \lrasseg---a model analog of the physical act of ``poking'' multiple times at a location in a static image and observing, in the short (imagined) videos that would result, what else in the scene is likely to move in response. To execute this procedure, we must first determine which regions in the image are capable of moving when acted upon by external forces. For example, a chair or car might respond to a push, while static elements like the sky or ground would not. Once these candidate regions are identified, we simulate a virtual poke at a specific pixel and ask the model: what else in the scene would move as a result? To realize this query, we extract two intermediate representations from \lrasseg: (1) a motion affordance map, indicating regions that are likely to move when external forces are applied, and (2) an expected displacement map, predicting how the rest of the scene would move in response to a virtual poke. We sample poke locations from high-affordance regions (as shown in Figure~\ref{fig:teaser}) and analyze resulting displacement fields from diverse virtual pokes to estimate patterns of correlated motion---yielding partitions of pixels (or segments) that describe co-moving entities. This statistical counterfactual probing framework not only enables object discovery but also holds promise for general-purpose robotics, where understanding movable parts of an environment and the distribution of likely responses to hypothetical applied forces is essential for planning and control. On our \spelkeentity benchmark, \lrasseg outperforms both self-supervised approaches like DINO~\cite{oquab2023dinov2} and supervised methods like SAM~\cite{ravi2024sam2}.

We then explore how these discovered Spelke objects can be \textit{used} in practice, focusing on the task of 3D physical object manipulation. In current practice, the typical pipeline for 3D object manipulation begins by identifying an object to manipulate using an off-the-shelf segmentation model, followed by applying a transformation to the selected segment to produce an edited image. While various techniques exist for performing these edits, most assume a predefined segment that corresponds to an object in the physical world.  However, since standard segmentation models often reflect semantic categories or texture similarity (as in SAM), but not entities that move together, the edits can often appear unrealistic---affecting parts of multiple objects or fragmenting a single object. In contrast, as shown on the right side of Figure~\ref{fig:teaser}, we find that using segments extracted using \lrasseg leads to edits that are significantly more physically plausible and align better with human expectations. We evaluate \lrasseg on the 3DEditBench benchmark~\cite{lee20253d}, which assesses the realism and physical consistency of 3D object manipulations. Our approach outperforms supervised baselines, including SAM, demonstrating the importance of using segments defined based on what moves together in the world (i.e. Spelke segments) for tasks involving physical manipulation.

In this way, our work addresses three key questions: (a) How can we \textit{benchmark} segmentation models' ability to capture the notion of pixel co-movement (i.e Spelke segments)? (b) How can such segments be defined using a visual world model and \textit{discovered} in a self-supervised way? and (c) Are Spelke segments a practically useful definition, i.e. can they be \textit{used} effectively in downstream physical manipulation tasks?

\section{Related Works}
\textbf{Evaluation datasets for segmentation:} Segmentation in computer vision is typically framed as semantic~\cite{zhou2018semanticunderstandingscenesade20k, lin2015microsoftcococommonobjects}, instance~\cite{lin2015microsoftcococommonobjects, gupta2019lvis}, or panoptic segmentation~\cite{kirillov2019panoptic}, and segments are labeled by annotators based on category. They will search through the image for objects matching a given category description and label the segment accordingly. However, these definitions often conflict with the physical structure of scenes: masks may merge independently movable objects, split objects into parts, or include amorphous “stuff” categories like sky or terrain~\cite{Qi_2023_EntitySeg}. Models trained and evaluated on such datasets, including recent efforts like SAM~\cite{ravi2024sam2}, may produce segments that diverge from the demands of downstream tasks like interaction and control~\cite{kroemer2021review}. In contrast, our work introduces a new segmentation benchmark called \spelkeentity which is aligned with the notion of Spelke objects~\cite{spelke1990principles}---a category-agnostic definition of segments as bounded regions that move as physical units, which allows for a more physically grounded evaluation of segmentation models.

\textbf{Object segmentation models:} Numerous approaches to object segmentation based on supervised learning have been proposed in prior work~\cite{he2017mask, cheng2021maskformer, alex2023segment, zou2023segment, jain2023oneformer, lueddecke22_cvpr}. While these models achieve strong performance, they depend on large amounts of labeled data, which can be expensive to obtain. To reduce reliance on annotations, recent unsupervised and self-supervised methods attempt to extract object masks from unlabeled data by clustering attention maps from pre-trained contrastive learning models~\cite{caron2021emerging, siméoni2021localizingobjectsselfsupervisedtransformers, wang2023tokencutsegmentingobjectsimages, hamilton2022unsupervisedsemanticsegmentationdistilling, xu2022groupvitsemanticsegmentationemerges, vangansbeke2022discoveringobjectmaskstransformers}. Methods like CutLER~\cite{wang2023cut} and ProMerge~\cite{li2024promerge} use these clusters as pseudo-masks for distillation. However, these methods often struggle in scenes with multiple objects of the same category, as contrastive learning tends to produce similar representations for such instances, making them harder to distinguish. Here, we introduce \lrasseg, a self-supervised world model that extracts segments based on what moves together in the physical world. We show that this approach is well-suited for downstream tasks such as object manipulation, compared to other ways of defining segments. 

 
\textbf{Emergent visual structures in world models:} 
Object-centric world models aim to decompose scenes into discrete entities by imposing inductive biases that encourage low-dimensional, disentangled object-centric representations to emerge. These models route information through a fixed number of competing latent slots, often using soft attention bottlenecks and minimizing either future prediction objectives (models such as MONet~\cite{burgess2019monet}, Slot Attention~\cite{locatello2020object}, SAVi~\cite{kipf2021conditional}) or contrastive objectives like C-SWM~\cite{kipf2019contrastive}. These models do not scale to complex real-world datasets due to architectural constraints---too many slots lead to degenerate solutions with minimal structure, while too few make the reconstruction task ill-posed. In contrast, \lrasseg defines a predictive world model that enables the extraction of segments grounded in motion causality without having to bake in architectural constraints. 

\textit{Counterfactual World Models (CWM)}~\cite{bear2023unifyingmachinevisioncounterfactual, venkatesh2024understanding} are a class of world models that reveal object-level structure by prompting a regression-based video predictor with targeted \textit{interventions}. Specifically, CWM simulates object motion by copying an RGB patch from the input frame to a different location in an otherwise fully masked target frame, and asking the model to reconstruct the target frame. The optical flow between the predicted and original frames reveals sets of pixels that tend to move together. However, this RGB-based intervention has important limitations. The copied pixel values often fail to reflect how that region would appear if the object truly moved—due to changes in lighting, occlusion, etc.—resulting in degraded predictions. Moreover, as CWM models are deterministic, when multiple future motions are plausible (e.g. for articulated bodies like humans), they produce blurred reconstructions that might result in inaccurate segments. \lrasseg resolves both these issues by a) specifying interventions via sparse flows that more meaningfully indicate intended motion and b) using autoregressive generative modeling to estimate the marginals of and sample over the distribution of plausible futures, avoiding averaging artifacts and enabling more accurate computation of correlated motion statistics for object discovery. 

\textbf{Text-guided vision foundation models:} Diffusion-based generative models~\cite{rombach2022high} have shown impressive capabilities in text-guided image generation and editing, and have an implicit understanding of objects and causal relationships. However, reliance on iterative global denoising makes it hard to probe these models with localized physical interventions, which is required for answering questions about causality---such as the effects of force application and object interactions. Similarly, while vision-language models (VLMs) such as CLIP~\cite{radford2021learning} and BLIP~\cite{li2022blip} excel at grounding global semantics in images, text-based prompts have proven to be a sub-optimal control surface for fine-grained spatial reasoning~\cite{komanduri2025causalvlbench, chen2024cello}. \lrasseg leverages the \lras autoregressive modeling framework to provide a robust control surface for spatially localized prompting, allowing sparse flow interventions to be specified simply by appending a few flow tokens to the input sequence.

\textbf{Object manipulation:} 
Object manipulation involves applying transformations to objects to generate novel scenes, and is a core task in computer vision. Most modern methods rely on spatial segmentation masks to define the set of pixels to be edited~\cite{wu2024draganything, shi2024lightningdraglightningfastaccurate, pandey2024diffusionhandles, yang2025transferringfoundationmodelsgeneralizable, gu2025diffusionshader3dawarevideo}. For physically plausible edits, it is critical that these segments correspond to parts of the scene that move together in the real world. However, commonly used models like SAM~\cite{ravi2024sam2} often produce masks that capture subparts of objects that often do not move independently of the rest of the scene, leading to implausible image edits. In contrast, \lrasseg produces segments aligned with real-world physical motion, enabling more accurate and realistic object manipulations for a wide range of physical object manipulation models.

\section{Methods}
\subsection{Benchmarking Spelke segments: The SpelkeBench Benchmark}
\label{sec:spelke_benchmark}

To benchmark Spelke object discovery, we introduce the \spelkeentity benchmark—a curated set of 500 images with ground-truth Spelke segment annotations. These annotations follow the definition proposed by developmental psychologist Liz Spelke~\cite{spelke1990principles}---Spelke segments are groups of pixels that move together as a unit under a variety of virtual pokes applied to the object. Existing benchmarks like COCO~\cite{lin2015microsoftcococommonobjects} and ADE20K~\cite{zhou2018semanticunderstandingscenesade20k} prioritize semantic or instance-level distinctions, often producing segments that merge independently movable objects, split them into parts, or include amorphous background regions. As illustrated in Figure~\ref{fig:ours_vs_SAM}, models like SAM frequently produce segments that diverge from Spelke criteria. Since such models are evaluated using existing benchmarks, it is hard to quantify their utility for tasks like robotic manipulation~\cite{kroemer2021review}, which require an understanding of which parts of a scene move together in the physical world. To address this gap, we introduce a method to construct a meaningful benchmark that tests whether models understand the pixel co-movement/Spelke object concept.

We curate a dataset of segmented objects from two complementary sources: the EntitySeg benchmark~\cite{Qi_2023_EntitySeg} and the OpenX-Embodiment robotics dataset~\cite{open_x_embodiment_rt_x_2023}. These datasets differ in their collection paradigms: EntitySeg is designed for high-resolution internet imagery with dense segmentation annotations, whereas OpenX consists of real-world, egocentric robot interactions. This contrast allows us to evaluate segmentation models in both unconstrained image domains and physically grounded robotics environments.

Since OpenX does not provide segment labels, we manually annotate \textit{Spelke-consistent segments} for a subset of 50 images. These annotations reflect the types of objects relevant for physical interaction and manipulation tasks that are central to robot learning. For EntitySeg, we extract a high-quality subset of 500 images using a three-stage filtering pipeline to filter out the annotated segments in the dataset which do not align with Spelke’s principles:

\begin{itemize}
    \item \textit{Stage 1: Removal of amorphous background regions.} We exclude all regions labeled as ``stuff''—such as sky, ground, or terrain—based on the standard stuff-vs-things taxonomy~\cite{Qi_2023_EntitySeg}. These regions lack the individuated, cohesive properties associated with Spelke objects and are typically not physically manipulable entities.

    \item \textit{Stage 2: Filtering non-movable object categories.} Despite being labeled as ``things'', certain objects like kitchen sinks, traffic signs, or large fixtures are functionally immovable in real-world settings. We identify and remove such regions through manual inspection.

    \item \textit{Stage 3: Final curation of diverse, high-quality scenes.} From the filtered pool, we select 500 images that contain only Spelke-consistent regions. We also ensure that this set is diverse in terms of object types, spatial arrangements, and scene complexity.
\end{itemize}


\subsection{Discovering Spelke segments}

\textbf{Local Random Access Sequence Modeling }(\lras)~\cite{lee20253d}, is a sequence modeling framework inspired by large language models (LLMs) that causally predicts locally quantized image (i.e. RGB) and optical flow patches. In this section, we describe the \lras architecture and provide details about how some of its properties make it a strong candidate for our goal of Spelke object discovery.

The \lras framework operates on a unified vocabulary comprising RGB and flow ``content'' tokens and a set of ``pointer'' tokens for each modality that specifies one of \( l \) spatial locations in the image grid---resulting in a vocabulary \( \mathcal{V} \) that can be partitioned into four disjoint sets of integers:
\begin{itemize}
    \item \( \mathcal{I}^{\text{(rgb)}} \): \text{RGB pointer} tokens — \( [0, l) \)
    \item \( \mathcal{X} \): \text{RGB content} tokens — \( [l, l + |\mathcal{X}|) \)
    \item \( \mathcal{I}^{\text{(flow)}} \): \text{Flow pointer} tokens — \( [l + |\mathcal{X}|, 2l + |\mathcal{X}|) \)
    \item \( \mathcal{F} \): \text{Flow content} tokens — \( [2l + |\mathcal{X}|, 2l + |\mathcal{X}| + |\mathcal{F}|) \)
\end{itemize}

\begin{figure*}[t]
  \centering
  \includegraphics[width=\textwidth]{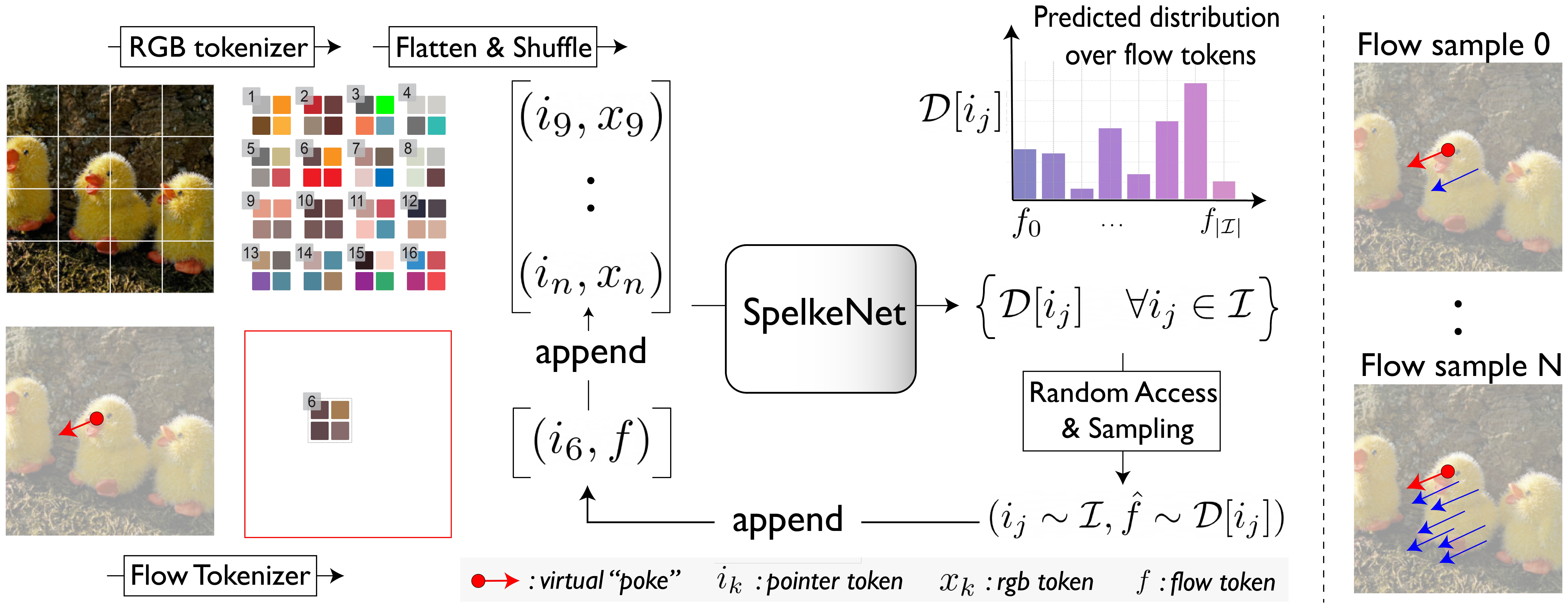}
  \vspace{-0.5em}
  \caption{\textbf{\lrasseg Architecture.} The \textbf{left} panel illustrates \lrasseg---an instance of the \lras~\cite{lee20253d} framework applied to the task of optical flow completion for Spelke object discovery. The input consists of a tokenized RGB image ($\{x_k, \, k \in \mathcal{I}\}$) and a sparse \textit{virtual poke} indicated by a flow token, $f$. Each token is paired with a pointer token indicating its spatial location, forming a 1D sequence of (pointer, content) pairs. The model accepts this sequence and predicts a categorical distribution $\mathcal{D}[i_j]$ over the flow token vocabulary for \textit{every} spatial location $i_j$ in the image. The \textbf{right} panel shows that autoregressively sampling from these distributions yields a complete flow field in pixel space---at each step we randomly select an undecoded location $i_j$, and sample a flow token $\hat{f} \sim \mathcal{D}[i_j]$ from the distribution predicted by the model. We then append the pair $(i_j, \hat{f})$ to the input sequence, which is fed back into the model to generate a new distribution $\mathcal{D}$, and the process repeats. In this way, the input sequence grows over time, progressively completing the flow field, representing how the scene responds to the virtual poke. We discover Spelke segments by analyzing the motion correlation patterns of these resulting flow fields.}
  \label{fig:main}
  \vspace{-1.0em}
\end{figure*}

When constructing sequences, each \text{content} token (i.e. RGB or flow) is paired with a corresponding \text{pointer} token that specifies its spatial location. This \text{(pointer, content)} pairing allows sequences to be arranged in arbitrary spatial order. Additionally, since the pointer tokens are modality-specific, they serve as a way of ``asking'' the model to decode a desired modality. For example, a pointer token from \( \mathcal{I}^{\text{(rgb)}} \) prompts the model to decode an RGB token at a given location, while one from \( \mathcal{I}^{\text{(flow)}} \) can query for a flow token at that same location. Token sequences are denoted as\footnote{In practice, for efficiency purposes, we reduce the number of pointer tokens by grouping each pointer with a patch of content tokens---each pointer token is followed by four content tokens as illustrated in Figure~\ref{fig:main}.}
\[
\mathbf{x} = [(i_1^{(\text{rgb})}, x_1), \dots, (i_N^{(\text{rgb})}, x_N)], \quad 
\mathbf{f} = [(i_1^{(\text{flow})}, f_1), \dots, (i_M^{(\text{flow})}, f_M)]
\]
\[
x_t \in \mathcal{X}, \quad f_t \in \mathcal{F}, \quad  
i_t^{(\text{rgb})} \in \mathcal{I}^{(\text{rgb})}, \quad 
i_t^{(\text{flow})} \in \mathcal{I}^{(\text{flow})}
\]

A special camera pose token $c$ representing the relative camera motion between frames can optionally be included to form the final sequence, \(\mathbf{z} = \mathbf{x} \oplus [c] \oplus \mathbf{f} \). Here, $\oplus$ denotes concatenation. 

The model is trained like an LLM---it learns to predict the next token, conditioned on all preceding tokens. More concretely, the model outputs a categorical distribution over the unified vocabulary $\mathcal{V}$, and is trained to minimize cross-entropy loss between this distribution and the target next token. Since the tokens appear in random spatial order, there is no need to ``learn'' the ordering, so the prediction of the pointer tokens is not supervised~\cite{lee20253d}. 

During inference, the sequence model can accept a sequence composed of any subset of the combined sequence $\mathbf{z}$. In prior work~\cite{lee20253d}, the \lras framework was used to generate a complete flow field by conditioning only on RGB tokens and the camera pose (i.e. \(\mathbf{z} = \mathbf{x} \oplus [c]\)), achieving state-of-the-art performance on tasks like monocular depth estimation and novel view synthesis~\cite{lee20253d}. 

\textbf{\lrasseg: an instance of \lras for Spelke object discovery.} Here, our goal is to discover pixel co-movement in natural images, i.e. what moves together when external forces are applied. We propose to discover such causal relationships by injecting localized virtual pokes and using a world model to infer what else in the scene moves. Among existing generative world modeling techniques, the \lras paradigm is particularly well-suited for this task. Unlike diffusion models~\cite{rombach2022high}, which require dense, global conditioning, the autoregressive structure of \lras supports composable input sequences. Our model, \lrasseg is a specific instance of \lras that leverages it's flexible sequence design properties to apply sparse, localized interventions simply by appending to the input sequence a flow token \(f_k\), representing the motion to be applied and a pointer token \(i_k\), indicating the spatial location of the poke---and discovers Spelke objects by completing the flow field which indicates how the rest of the scene will respond to the poke. 

\label{sec:seq_design}

\textit{Disentangling object motion from camera motion.} In natural videos, pixel motion can arise either due to external forces acting on objects or due to camera movement. However, for discovering Spelke objects, we are specifically interested in motion caused by external perturbations to objects, not from camera-induced motion. If we provide the model with a sparse flow input (a virtual poke) without additional context, it has no way of knowing whether the input motion arose from forces acting on objects or camera movement. Consequently, the model might complete the flow field in ways that conflate both sources of motion, making it difficult to isolate responses that are effects of the applied poke. To ensure that the predicted motion is attributed solely to the virtual poke, we must explicitly condition the model on a \textit{static camera}. With the \lras paradigm, this form of controlled probing is simple to implement: we simply append a zero camera pose token to the input sequence, guiding the model to interpret any input motion as arising exclusively from external forces and not from camera displacement. We formalize this input as:
\[
\mathbf{z}_f = \mathbf{x} \oplus [c = 0] \oplus [(i_k, f_k)],
\]

\textbf{Decoding strategies.} Given the input sequence, \lrasseg predicts for every spatial location, $i_k \in \mathcal{I^{\text{(flow)}}}$, a categorical distribution $\mathcal{D}[i_k]$ over $\mathcal{V}$.\footnote{Although the model predicts a distribution over the entire unified vocabulary, not just the flow token subset \( \mathcal{F} \), sampling from this distribution yields a flow token \( f \sim \mathcal{D}[i_k] \) because the model is trained to produce a flow token whenever it encounters a preceding flow pointer token.} In practice, $\mathcal{D}[i_k]$ is obtained by querying the model by appending a pointer token to the end of the sequence:
\[
\Psi(\mathbf{z}_f \oplus [i_k]) \mapsto \mathcal{D}[i_k]
\]

\lrasseg can be thought of as a composite function \(\Psi\) that returns a set of flow distributions, one for each spatial location:
\[
\Psi(\mathbf{z}_f) = \left\{ \mathcal{D}[i_k] \,\, \forall \, i_k \in \mathcal{I}^{\text{(flow)}} \right\}
\]

To infer what else in the scene will move as a result of the poke $f_k$, we can use $\Psi$ to complete the flow field, either in parallel or autoregressively. In parallel decoding, all spatial locations are sampled independently, $\hat{f}_k^{\text{(par)}} \sim \mathcal{D}[i_k]$, resulting in the spatially completed flow field, $\hat{\mathbf{f}}^{\text{(par)}} = [\hat{f}_k^{\text{(par)}}]_{k \in \mathcal{I}} $. We denote this method as $\Psi_{\text{flow}}^{\text{par}}$:
\[
\hat{\mathbf{f}}^{\text{(par)}} = \Psi_{\text{flow}}^{\text{par}}(\mathbf{z}_f).
\]

In contrast, sequential decoding starts from an initial sequence $\mathbf{z}_0 = \mathbf{z}_f$. We then iteratively select an undecoded, random location $i_k$, query the current model distribution $\mathcal{D}_0 = \Psi(\mathbf{z}_0)$, sample a token: \( \hat{f}_k^{\text{(seq)}} \sim \mathcal{D}_0[i_k], \)
and append it to the sequence:
\[
\mathbf{z}_{q+1} \leftarrow \mathbf{z}_q \oplus \{(i_k, \hat{f}_k^{\text{(seq)}})\}
\]
This process continues until the entire flow field is decoded, with each prediction step refining the model’s estimate of distribution: $\mathcal{D}_{q+1} = \Psi (z_{q+1})$. This sequential decoding process results in the dense flow field, $\hat{\mathbf{f}}^{\text{(seq)}} = [\hat{f}_k^{\text{(seq)}}]_{k \in \mathcal{I}}$. We denote this method as $\Psi_{\text{flow}}^{\text{seq}}$:
\[
\hat{\mathbf{f}}^{\text{(seq)}} = \Psi_{\text{flow}}^{\text{seq}}(\mathbf{z}_f; \texttt{seed}=t)
\]

Sequential decoding is especially valuable when modeling objects with many degrees of freedom, such as articulated bodies like humans or mechanical tools, and deformable materials like cloth or paper, where different parts are causally linked and must move in a coordinated way. For example, in human motion, movement of one part—say, the lower hand—imposes constraints on how other parts, like the upper arm or torso, can respond. Decoding tokens sequentially allows the model to respect these causal dependencies, as each token is generated in the context of previously decoded ones (such as the motion of the lower hand in this example), resulting in globally consistent motion. In contrast, parallel decoding offers faster inference but can yield locally plausible yet globally inconsistent flow fields---e.g. the upper arm moving independently of the lower hand—leading to physically implausible outcomes.

\textbf{Defining Spelke objects using \lrasseg.}  Our approach builds on the idea of \emph{counterfactual probing}, introduced in CWM~\cite{venkatesh2024understanding}, where Spelke objects are discovered by simulating localized virtual pokes through local patch motion interventions and analyzing the outcome of the intervention. However, since CWM is regression-based, it produces a single deterministic output---an important limitation, because in the physical world, responses to pokes are often \textit{multimodal}. Consider a simple example of moving a person's hand. In the physical world, one of two things can plausibly happen: either the hand moves independently while the rest of the body remains fixed, or the entire person translates, causing the hand to move along with it. Both are physically valid outcomes. But as CWM is deterministic, it is forced to average over these distinct possibilities, leading to blurry or ambiguous motion completions that fail to reveal which parts of the scene tend to move together.

To address this limitation, we propose a more expressive definition of Spelke objects using a generative world model like \lrasseg, which generates multiple plausible future motions of a scene. We operationalize Spelke objects as groups of pixels that consistently move together across multiple plausible outcomes of a world model, under different virtual pokes. This requires modeling the \textit{distribution} of possible responses to external forces.

We implement this using \emph{statistical counterfactual probing} on \lrasseg, a stochastic extension of the original CWM counterfactual procedure. Instead of generating a single prediction like CWM, we use \lrasseg to produce a diverse set of \textit{imagined} flow completions for various virtual pokes at a candidate spatial location. Diversity arises from two sources of randomness:

\begin{enumerate}
    \item \textbf{Sampling flow tokens} from the learned distribution $\mathcal{D}[i_k]$: For a fixed index $i_k$, we draw multiple flows $f_k \sim \mathcal{D}[i_k]$ to explore the local responses the model deems plausible. For example, in the human motion scenario discussed above, different samples can make the same body part move in distinct yet physically feasible ways.
    \item \textbf{Varying the decoding order} of spatial indices $i_k$: Because \lrasseg is a sequence model, tokens decoded earlier condition those decoded later. Shuffling the order therefore changes how motion propagates through the object---e.g. decoding the torso \emph{before} the leg yields a different global outcome than decoding the leg first.
\end{enumerate}

Computing what is effectively a marginal (i.e. a probability-weighted integral) over these diverse generations, allows us to obtain a robust definition of Spelke objects using \lrasseg. We now describe a few useful structure extractions from \lrasseg that support statistical counterfactual probes.

\label{p_motion}
\textbf{Motion affordance maps}. To discover Spelke objects, we must first identify the candidate locations where virtual pokes can be applied---which pixels in the scene lie on regions that are likely to move under external forces (i.e. \textit{movable entities}).  We refer to this notion as the probability of motion affordance map, denoted \( p_{\text{motion}} \). Such motion-centric affordance maps are especially useful in robotics applications where we need to identify high motion affordance regions that are likely to move under interaction (e.g. a cup or plate). Regions that typically do not move upon external forces (e.g. sky, walls, and flooring) would have low motion affordance.

\begin{figure}[t]
  \centering
  \includegraphics[width=\textwidth]{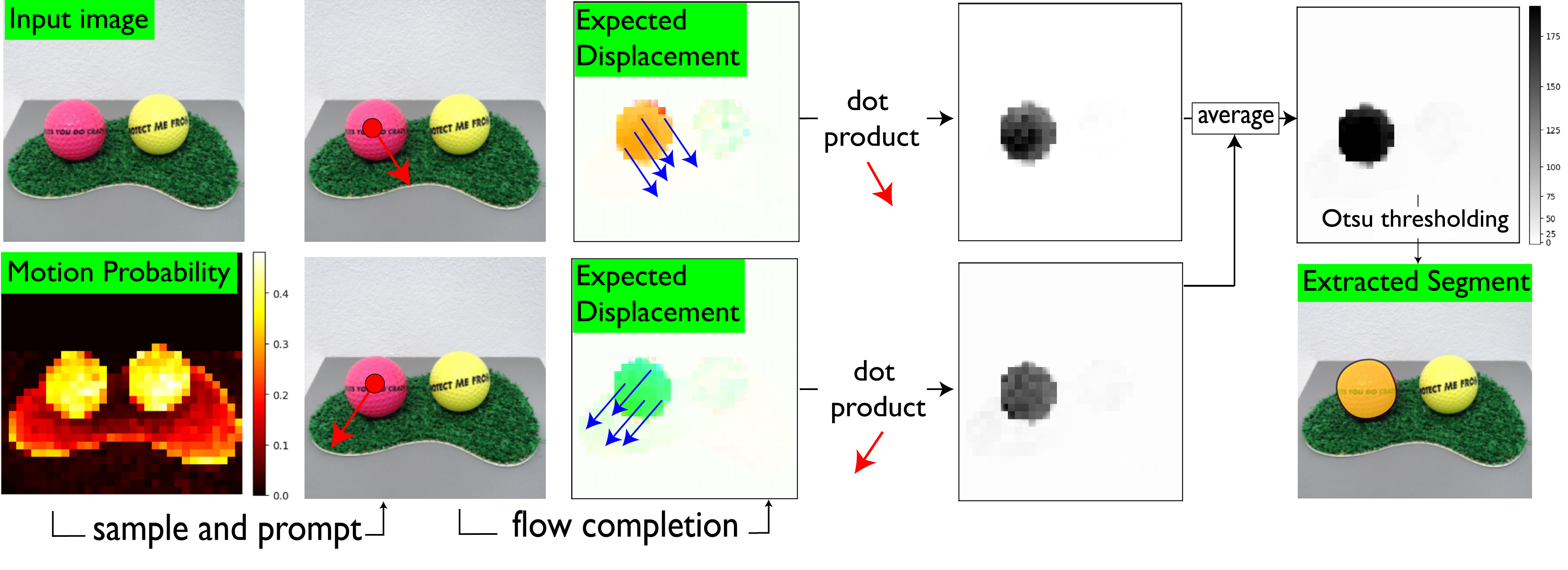}
  \vspace{-1em}
  \caption{\textbf{Spelke object discovery using statistical counterfactual probing}. To discover movable objects, multiple virtual pokes are applied at a location sampled on the $p_{\text{motion}}$ map that indicates which regions are likely to move under the application of external forces. The average dot product of the poke vector with the expected displacement maps isolates the desired segment.}
  \label{fig:spelke_discovery}
  \vspace{-1.0em}
\end{figure}

To compute \( p_{\text{motion}} \), we define a set of flow tokens that correspond to motion greater than some threshold $\tau$, and then sum their estimated probabilities. As the flow tokens, $f_j$, are by themselves not interpretable, we map each flow token to a 2D flow vector, $\mathbf{v}_j$, through an epigraphy on the flow vocabulary\footnote{\label{fn:epigraphy}Flow token epigraphy: \lrasseg uses a learnt \emph{local patch quantization} to produce flow tokens, but relies on a \emph{global decoder} to generate coherent, high-quality flow fields. As a result, tokens cannot be interpreted by decoding them in isolation---their meaning emerges only in the context of the full sequence. However, since the tokenizer is local, we can find which continuous flow vectors map to it by performing a kind of token space epigraphy---by assigning meaning to discrete flow tokens through statistical aggregation of typical input flow fields that produced them:
\[
f_j \!\mapsto\!  \mathbf{v}_j = \frac{1}{|S_j|} \sum_{\mathbf{u} \in S_j} \mathbf{u}, \quad \text{where } S_j = \left\{ \mathbf{u} \in \mathbb{R}^2 \mid \text{tokenizer}(\mathbf{u}) = f_j \right\}.
\]
} and define the token set corresponding to motion as:
\[
\mathcal{F}_{\text{motion}} = \left\{ f_j \in \mathcal{F} \,\middle|\, \|\mathbf{v}_j\|_2 > \tau \right\},  \text{where} \, \tau \, \text{is a threshold} 
\]
Next, given a sequence of RGB tokens $\mathbf{x}$, as we are only interested in finding regions that are likely to move under external forces, we concatenate the sequence with a token indicating zero camera motion to discount it (i.e. \( \mathbf{z} = \mathbf{x} \, \oplus \, [c=0] \)) and obtain the predicted flow token distributions \( \mathcal{D}[i_k] = \Psi(\mathbf{z})[i_k] \),  \( \forall i_k \in \mathcal{I} \). Using these distributions, the probability of \emph{motion} at each spatial location $i_k$, is computed by summing over the token set $\mathcal{F}_{\text{motion}}$:
\[
p_{\text{motion}}[i_k] = \sum_{f_j \in \mathcal{F}_{\text{motion}}} \mathcal{D}[i_k, j]
\]
In this way, \( p_{\text{motion}}: \mathcal{I} \rightarrow [0, 1] \) is a 2D heatmap of the regions likely to move under external forces. Figures~\ref{fig:spelke_discovery} and~\ref{fig:teaser} illustrate some examples of these maps. 


\textbf{Expected Displacement Maps.} Having identified regions of high motion-affordance, we can sample candidate locations and condition the model on virtual pokes in those regions. To discover Spelke objects, we introduce a useful quantity called the ``expected displacement map'', which is the estimate of the likely flow at each location conditioned on the poke. In robotics settings, this map can provide valuable guidance about how objects might move if interacted with, even before the robot makes contact with objects in the scene. 

We use our flow model to apply a virtual poke represented with the flow token, $f_k$, at location $i_k \in \mathcal{I}$, construct an input sequence,  \( \mathbf{z} = \mathbf{x} \oplus [c=0] \oplus [(i_k, f_k)] \), and obtain the predicted distribution, \( \mathcal{D}[i_k] = \Psi (\mathbf{z})[i_k] \),  \( \forall k \in \mathcal{I} \). We then compute the expected displacement as the probability-weighted average of flow vectors \( \mathbf{v}_j \), where each \( \mathbf{v}_j \) maps to token \( f_j \) as defined by flow token epigraphy\footref{fn:epigraphy}:
\[
\mathbb{E}_{\textrm{disp}}^{\textrm{par}}[i_k] = \sum_j \mathcal{D}[i_k, j] \cdot \mathbf{v}_j, \quad  \forall i_k \in \mathcal{I}
\]
The result is a dense 2D vector field over spatial locations, decoded in parallel: \(
\mathbb{E}_{\textrm{disp}}^{\textrm{par}}: \mathcal{I} \rightarrow \mathbb{R}^2 \). We denote this method using the function \( \Psi_{\text{disp}}^{\text{par}}\):
\[ \mathbb{E}_{\textrm{disp}}^{\textrm{par}} = \Psi_{\text{disp}}^{\text{par}}(\mathbf{z})
\]
To obtain a more faithful estimate of \( \mathbb{E}_{\textrm{disp}}^{\textrm{par}} \), we can also average predictions over multiple stochastic generations (i.e rollouts) of the model. Denoting $\hat{\mathbf{f}_t}^{\text{(seq)}} = \Psi_{\text{flow}}^{\text{seq}} (\mathbf{z}, \texttt{seed} =t)$ as the set of sequentially decoded flows in the \( t^\text{th} \) rollout, where each rollout is one sequential completion of the flow field conditioned on $\mathbf{z}$, the expected displacement map, computed in sequential mode can be written as:
\[
\mathbb{E}_{\text{disp}}^{\text{seq}}[i_k] = \frac{1}{T} \sum_{t=1}^T  \hat{\mathbf{f}_t}^{\text{(seq)}}[i_k] \cdot \mathbf{v}_j.
\]
We denote this method using the function \( \Psi_{\text{disp}}^{\text{seq}}\):

\[ \mathbb{E}_{\textrm{disp}}^{\textrm{seq}} = \Psi_{\text{disp}}^{\text{seq}}(\mathbf{z})
\]
Some examples of these maps are shown in Figure~\ref{fig:spelke_discovery}. To simplify the notation going forward, unless the superscript (seq/par) is specified, we'll assume either sequential or parallel modes can be used.

\textbf{Statistical counterfactual probing for Spelke object discovery.} 
\label{sec:obj_discovery}
Using these structure extractions, we first sample a location that is likely to move: \( k\) such that \( p_{\text{motion}}(k) > \tau_p \). Then, to discover Spelke objects we will identify regions that consistently move together under various virtual pokes applied at \( k \). We use our flow model to apply diverse virtual pokes \( \{ f^{(r)} \}_{r=1}^R \), at \( k \). For each direction \( f^{(r)} \), we compute the expected displacement field, given the input sequence:
\[
\mathbb{E}_{\text{disp}}^{(r)} = \Psi_{\text{disp}} (\mathbf{x} \oplus c=0 \oplus [i_k, f^{(r)}]).
\]

To discover co-moving entities (i.e. Spelke objects), we computed the \textit{expected motion correlation}, $\bar{\text{dot}}[u]$ at each location \( u \in \mathcal{I} \), by averaging across various pokes, the dot product between the poke vector \( f^{(r)} \) and the expected displacement map $\mathbb{E}_{\text{disp}}^{(r)}[u]$:

\[
\bar{\text{dot}}[u] =  \frac{1}{R} \sum_{r=1}^R \left\langle f^{(r)}, \mathbb{E}_{\text{disp}}^{(r)}[u] \right\rangle.
\]

Finally, Otsu thresholding \cite{4310076} of \( \bar{\text{dot}} \) yields our desired Spelke segment. Refer to Figure~\ref{fig:spelke_discovery} for a more detailed illustration of this procedure. In practice, we find that using the sequential model to aggregate over multiple stochastic generations of the model is more effective. This is especially the case for objects with many degrees of freedom, like humans and deformable objects (see above), although more expensive to compute. However, we also find that reasonable results can be achieved in parallel mode as well.

\label{sec:autoseg}
\textbf{Automatically discovering every Spelke object in a scene}. So far, we have shown how Spelke segments can be extracted from point prompts. However, in many real-world settings, especially in robotics, it is advantageous to automatically discover \textit{every} independently movable segment/Spelke object in a scene without requiring manual point-prompting. For example, a household robot tasked with clearing a dining table must infer that a plate and its contents will move as a unit, while a napkin resting on the plate is an independent entity, so it can plan appropriate grasps and avoid unintended collisions. 

We now describe a method to extract the full set of Spelke segments in a scene automatically. Our approach consists of two steps. First, we compute a dense pixel-to-pixel affinity matrix that captures the likelihood that a pair of pixels will move together under virtual force. In essence, this process recovers the pairwise causal structure of the scene, revealing which regions are causally entangled in motion space. An iterative clustering algorithm is then applied to this matrix to isolate a complete set of independently movable entities. 

\begin{figure}[t]
  \centering
  \includegraphics[width=\textwidth]{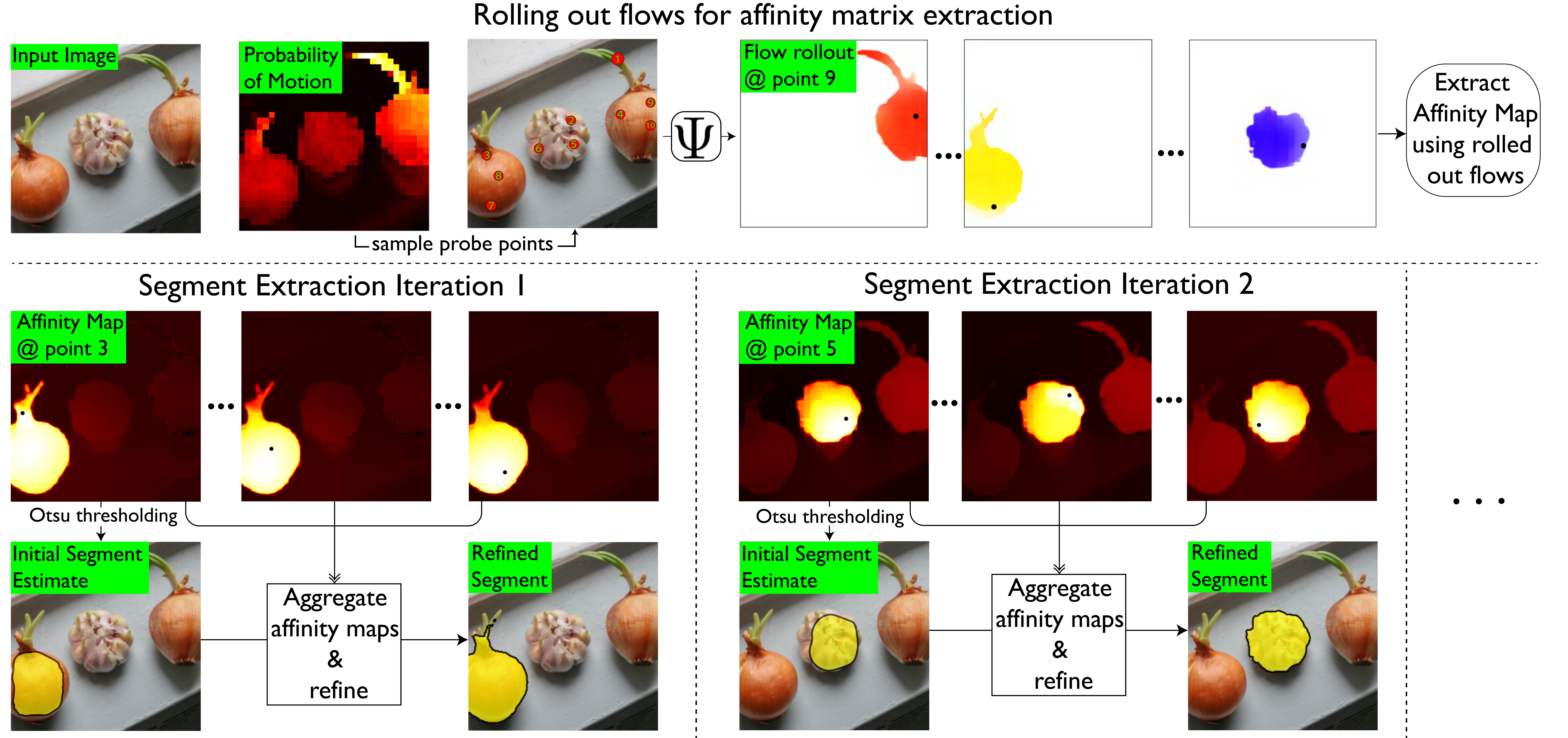}
  \vspace{-1em}
  \caption{\textbf{Automatic discovery of Spelke segments}. We extract probability of motion maps from an image, and use it to sample candidate poke points \textbf{(top left)}. We apply an optical flow vector ``poke'' to the image at the sampled points and obtain dense flow fields conditioned on the poke \textbf{(top right)} which are used to compute affinity maps. As shown in the \textbf{bottom} panel, these maps enable the extraction of segments using iterative clustering (see Section~\ref{sec:autoseg}).}
  \label{fig:autosegschematic}
  \vspace{-1.0em}
\end{figure}

\textit{Computing the affinity matrix}. We begin by sampling locations from the motion affordance map. These points are where we “poke” to collect flows.
\[
\mathcal{K} \;=\;\{\,k_{1},\,k_{2},\,\dots,\,k_{N}\}
\;\subset\;\mathcal{I},
\qquad
p_{\mathrm{motion}}[k_i]\;>\;\tau_{p}.
\]

We then build a motion descriptor for each pixel using the following procedure:

For each \(n=1,\dots,N\), choose \(R\) poke‐directions \(\{f_n^{(r)}\}_{r=1}^R\).  For each \((n,r)\) and each of \(t=1,\dots,T\) random seeds, compute the flow completion given the input image tokens $\mathbf{x}$,
\[
\hat{\mathbf{f}_t}^{(n, r)} = \Psi_{\text{flow}}^{\text{seq}} (\mathbf{x},c=0, [i_{k_n}, f_n^{(r)}], \texttt{seed}=t)
\]
Then for each \(u\in\mathcal{I}\) the motion descriptor,
\[
\varphi[u] =\bigl[\hat{\mathbf{f}_1}^{(1, 1)},\,\dots,\,\hat{\mathbf{f}_t}^{(n, r)}(u)\bigr]
\;\in\;\mathbb{R}^{2\,N\,R\,T}.
\]
Finally, the affinity matrix can be described as the pairwise dot product of motion descriptors: 
\[
A[u,v]=\varphi[u]^\top\,\varphi[v],
\quad
\forall\,u,v\in\mathcal{I}.
\]
For simplicity, we denote $A[u]$ to be the affinity of the pixel $u$ with the rest of the image. 

\textit{Clustering the affinity matrix to extract segments.}  Given the precomputed affinity matrix \(A\), we extract segments in an iterative “select–threshold–refine” loop. At each step, we choose the most confident probe center \(k_{i^*}\), defined as the one whose affinity‐row \(A[k_{i^*}]\) has the highest mean over all pixels---indicative of strong binding to the other pixels that make up the object. We apply Otsu’s method to threshold this row, yielding an initial mask \(M^{(0)}\). We then gather all remaining poke points \(k_j\) that lie within \(M^{(0)}\) and average their affinity‐rows to form:
\[
A_{\mathrm{avg}}
= \frac{1}{\lvert\{j: k_j\in M^{(0)}\}\rvert}
  \sum_{k_j \in M^{(0)}} A[k_j]
\]
We threshold \(A_{\mathrm{avg}}\) via Otsu's method to obtain a refined mask \(M ^ {(t)}\), for $t=0$. All centers contained in \(M^{(t)}\) are then removed from consideration, and the loop repeats on the remaining set of poke points. Once no poke points remain, the algorithm returns the complete set of extracted segments \(\{M^{(1)},\dots,M^{(T)}\}\). Figure~\ref{fig:autosegschematic} illustrates this procedure using an example. 

\subsection{Using Spelke segments for physically plausible object manipulation}

\begin{figure*}[t]
  \includegraphics[width=\textwidth]
  {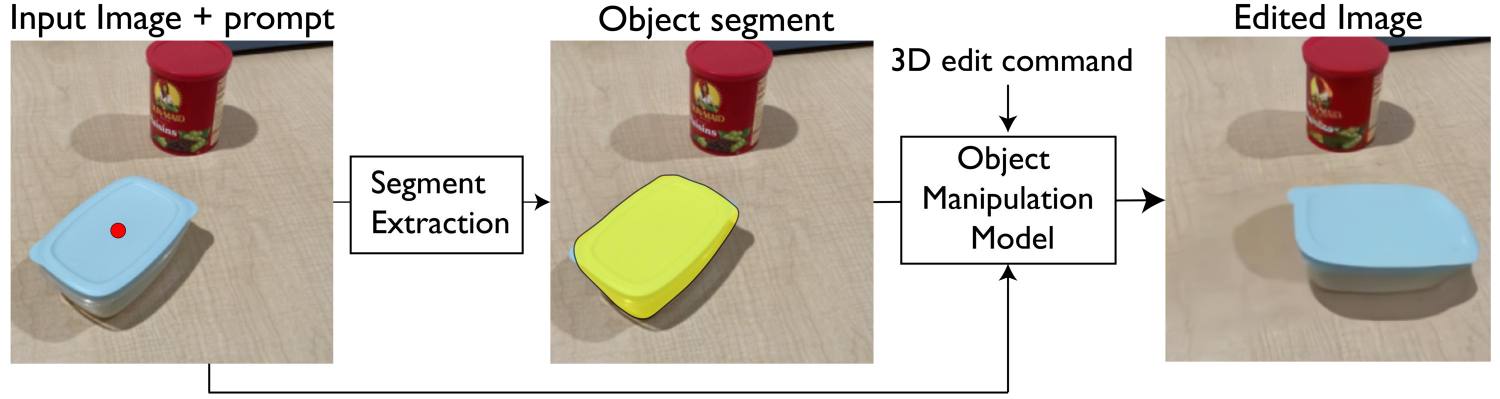}
  \vspace{-1em}
  \caption{\textbf{Standard pipeline for object editing using segmentation masks:}
 We substitute SAM segments for Spelke objects predicted by \lrasseg and find that they yield more intuitive and physically plausible edits.}
  \label{fig:obj_manip_schem}
\end{figure*}

Now that we have described how Spelke segments can be discovered given point samples, we discuss how they can be used in practical applications that require an understanding of pixel co-movement. We consider the standard task of object manipulation shown in Figure~\ref{fig:obj_manip_schem} : an input image is given along with a user-defined edit prompt specifying the desired 2D/3D transformation, and an object mask tells the editing model which parts of the scene to apply the transformation on. Successful manipulation relies on having a physically meaningful object segment as downstream object transformations can suffer if a segment corresponds to a region that is not independently movable. Spelke segments are grounded in physical principles: they group pixels based on correlated motion under virtual forces. This makes them a more suitable primitive for physically plausible editing---here, we demonstrate that the choice of segmentation method significantly affects the realism of the edit. In this way, we show that Spelke segments are not just a theoretical concept, but have practical utility in downstream tasks.

\paragraph{The \lras framework enables both image editing and segmentation via flexible sequence design.} In this paper, we leverage the \lras framework to build \lrasseg---a flow completion model for discovering Spelke segments. To recap, \lrasseg is trained to complete flow fields conditioned on an input sequence comprising RGB tokens $\mathbf{x}$ and a sparse virtual poke $\mathbf{f}$. The input sequence is denoted as:
\[
\mathbf{z} = \mathbf{x} \oplus \mathbf{f},
\]
The model predicts a complete flow field representing how the rest of the scene will move as a result of the poke. By analyzing the resulting flow field, we discover Spelke segments.

The same underlying \lras framework can also be used to define an image editing model in pixel space. Prior work~\cite{lee20253d} demonstrates this by building \lrasd---an instance of the \lras framework that is conditioned on input RGB tokens $\mathbf{x}$ and dense flow tokens $\mathbf{f}^{\text{dense}}$ that specify a desired object transformation to be applied:
\[
\tilde{\mathbf{z}} = \mathbf{x} \oplus \mathbf{f}^{\text{dense}}.
\]
Given this sequence, the model predicts a distribution over RGB tokens:
\[
\mathcal{D}^{\text{rgb}}[i_k] = \Psi_{\text{edit}}(\tilde{\mathbf{z}})[i_k], \quad \forall i_k \in \mathcal{I}.
\]

To construct an $\mathbf{f}^{\text{dense}}$ that represents a transformation targeting a particular object in the scene, a segmentation mask must be specified. Prior work relies on off-the-shelf methods such as SAM~\cite{ravi2024sam2} to define these masks, which often represent regions which do not move as a unified whole, resulting in implausible edits.

Here, we show that the \lras framework itself can be used to define a segmentation model, removing the reliance on external segmentation methods. By a simple modification of the sequence structure—e.g., prompting with sparse pokes and analyzing the flow response—we obtain segments better aligned with the demands of physical reasoning and manipulation. In this way, we demonstrate in this paper that the \lras framework unifies both segmentation and image editing within a single, token-based autoregressive modeling paradigm.

\noindent

\section{Results}

\subsection{Point-prompted segmentation} 
\label{sec:point_prompted}

\textbf{Task \& Dataset.} To evaluate our model’s ability to extract Spelke objects, we formalize the task as point-promoted segmentation: given a point on an object, the goal is to recover the region that would move together if a virtual force were applied at that point. To evaluate Spelke object discovery, we use \spelkeentity: our 500-image benchmark described in Section~\ref{sec:spelke_benchmark}.

\textbf{Baselines.} 
We compare against several strong baselines. For supervised segmentation, we use SAM2 (heira-large)~\cite{ravi2024sam2}, a state-of-the-art point prompt-based method. For self-supervised baselines, we evaluate DINOv1~\cite{caron2021emerging} and DINOv2~\cite{oquab2023dinov2}, which reveal semantic object structure via attention maps---segments are obtained by thresholding attention maps at a prompted location. We also compare to Counterfactual World Models (CWM) \cite{bear2023unifyingmachinevisioncounterfactual, venkatesh2024understanding}, which segment objects by generating local patch motion interventions and thresholding the estimated optical flow (using RAFT~\cite{wang2024searaft}) between the outcome of the intervention and the original image.

\begin{figure}[b!]  
  \centering
  \includegraphics[width=\linewidth]{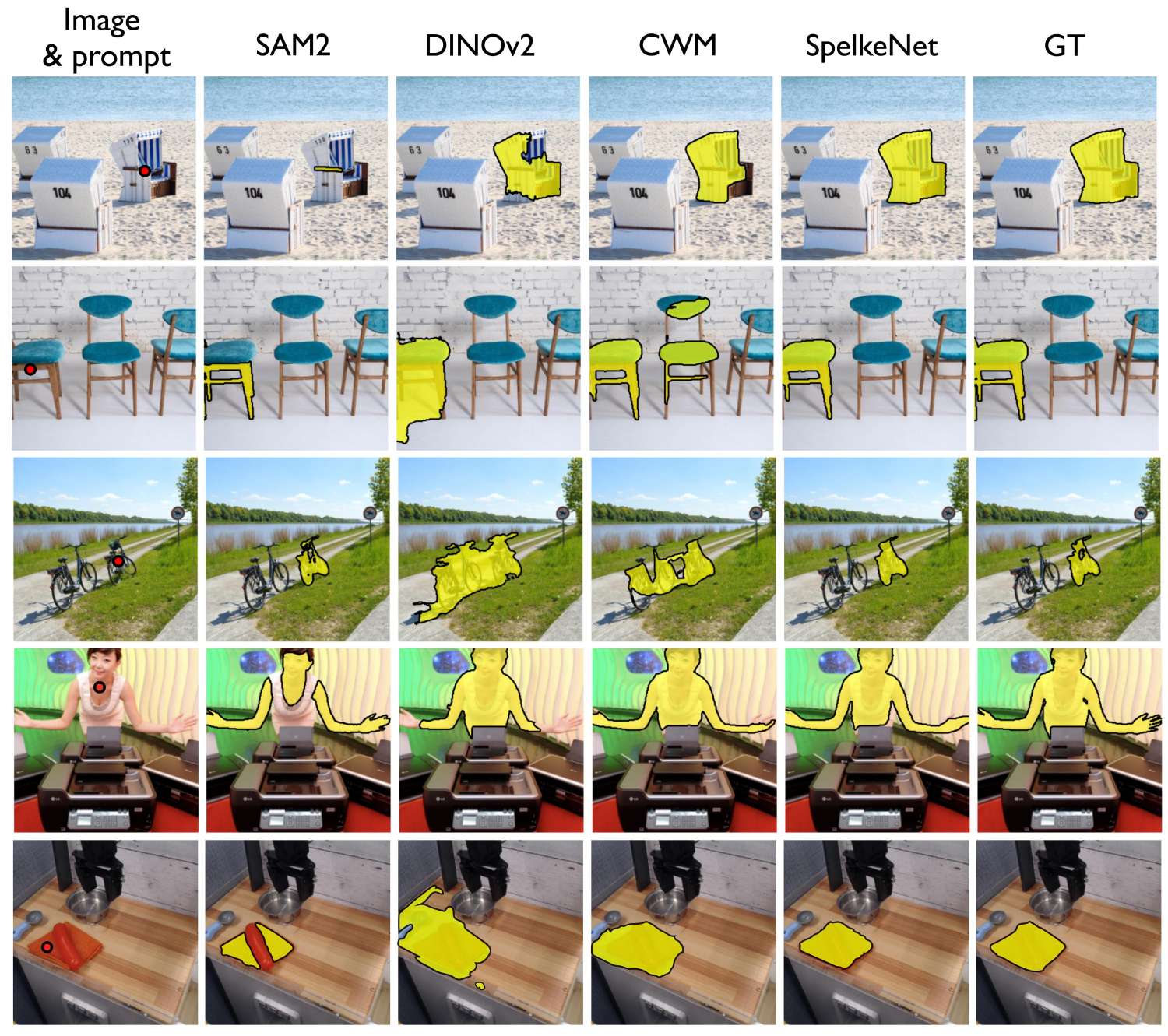}
  \vspace{-0.5em}
  \caption{\textbf{Qualitative results for point-promoted segmentation across models.} \lrasseg yields sharper segments, better aligned with Spelke's definition of grouping pixels based on co-movement, compared to SAM2, DINO, and CWM. More results are provided in the attached supplementry.}
  \label{fig:pointsegres}
\end{figure}

\begin{table}[t!]

  \centering
  \caption{\textbf{Quantitative evaluation of point-prompted segmentation accuracy across models.} We report Average Recall (AR) and mean Intersection over Union (mIoU) for various segmentation methods. \lrasseg outperforms both self-supervised baselines (DINO, CWM) and the supervised SAM2 model.}
  \label{tab:point_seg}
  \small
  \setlength{\tabcolsep}{4pt}
  \begin{tabular}{lcccccc}
    \toprule
      & SAM2 & DINOv1-B/8 & DINOv2-L/14
      & DINOv2-G/14 & CWM
      & \lrasseg \\
    \midrule
    AR    & 0.4816 & 0.2708 & 0.2524 & 0.2254 & 0.3271 & \textbf{0.5411}  \\
    mIoU  & 0.6225 & 0.4990 & 0.4931 & 0.4553 & 0.4807 & \textbf{0.6811}  \\
    \bottomrule
  \end{tabular}
\end{table}

\textbf{Evaluation details \& metrics.} 
For each ground truth segment, we generate a point prompt using the centroid or, if outside the mask, the point farthest from the boundary. We run 8 poke directions and 3 autoregressive flow completions per prompt with the procedure described in Section~\ref{sec:autoseg}, and we use the same setup for CWM. We use Average Recall (AR) and mean intersection-over-union (i.e. mIoU) to measure performance. AR is defined as the fraction of GT segments that the model detects. Here, a GT segment is classified as \textit{detected} if the predicted segment obtains an IoU less than some threshold $\tau$.  In practice, we compute the average AR across multiple IoU thresholds (0.5 - 0.99). Intuitively, we can think of AR as measuring how likely it is that the GT segments are detected by the model and the mIoU metric as measuring how precisely each segment boundary is predicted. 

\begin{figure}[b!]
  \centering
  \includegraphics[width=0.98\textwidth]{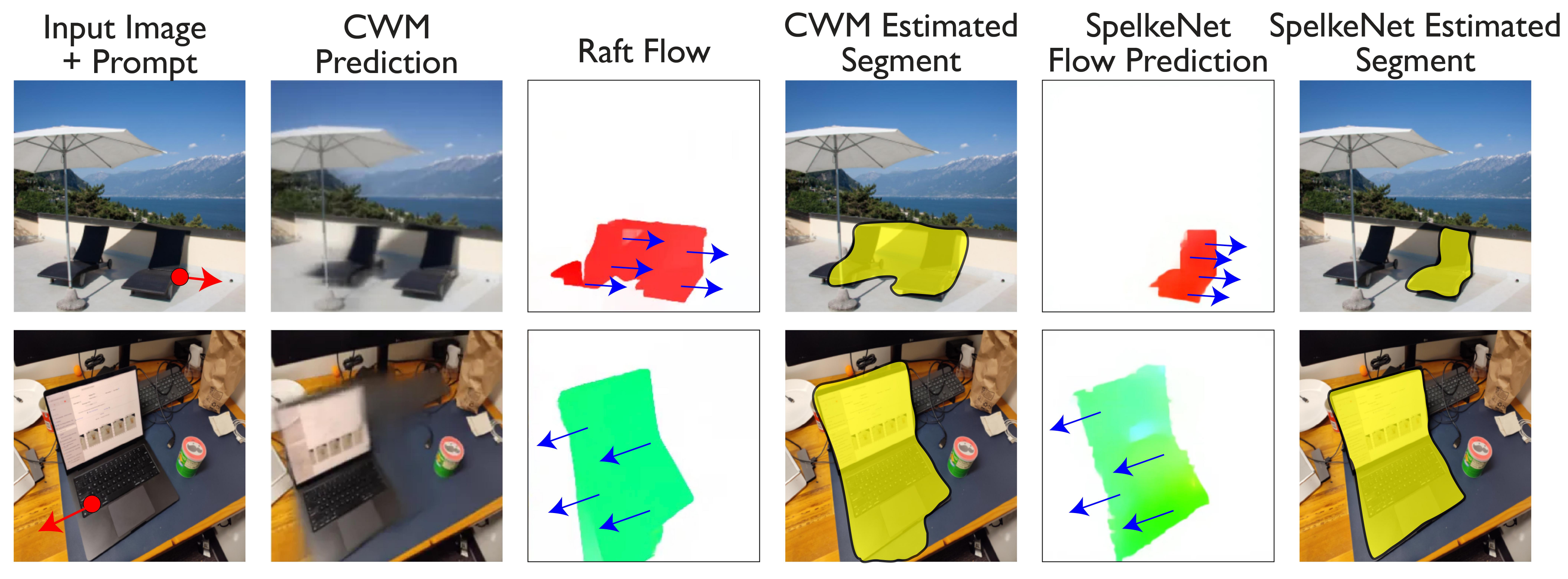}
  \vspace{1.0em}
  \caption{\textbf{CWM Segmentation failure modes in complex scenes.} Each row shows a challenging example where CWM struggles. The first column shows the input image with the patch motion prompt (red arrow). The second column displays the counterfactual prediction generated by CWM. The third column shows the RAFT-predicted flow field between the input and counterfactual image. The final column presents the resulting segment obtained by thresholding the flow magnitude. Compared to \lrasseg, CWM often produces diffuse motion fields due to blurry RGB reconstruction and inaccurate object boundaries.}
  \label{fig:cwmcomparisonsuppl}
\end{figure}


\textbf{Qualitative and Quantitative Comparisons}. \lrasseg outperforms all baselines across both Average Recall (AR) and mean IoU (mIoU), surpassing self-supervised methods like DINO and CWM, as well as the supervised baseline SAM, as shown in Table~\ref{tab:point_seg}.

Qualitatively, as depicted in Figure~\ref{fig:pointsegres}, although SAM performs well in many cases, it often segments non-movable regions such as textures, printed designs, or object subparts such as human skin, relying on appearance rather than physical coherence. This suggests that semantic and texture-driven segmentation can be misaligned with the goal of identifying physically grounded, movable objects. Meanwhile, contrastive learning methods like DINO exhibit a different failure mode, merging same-category instances, as the contrastive learning objective brings representations of instances of the same object closer. These observations highlight a fundamental limitation of such models for the task of discovering Spelke segments. CWM, while stronger than other self-supervised methods, often merges nearby objects. This happens because the model often generates blurry reconstructions, as its RGB pixel regression objective during training does not account for uncertainty. As a result, the flow estimation may produce diffuse or extended motion fields, causing nearby objects to be grouped together, as illustrated in Figure~\ref{fig:pointsegres} and ~\ref{fig:cwmcomparisonsuppl}.

In contrast, \lrasseg yields sharp, high-quality segments closely aligned with the Spelke definition. This can be largely attributed to the ability to prompt the model with local cues and to the probabilistic flow completion architecture that explicitly accounts for uncertainty in visual scenes.

\subsection{Automatic discovery of Spelke segments}

\begin{figure}[b!]
  \centering
  \includegraphics[width=\textwidth]{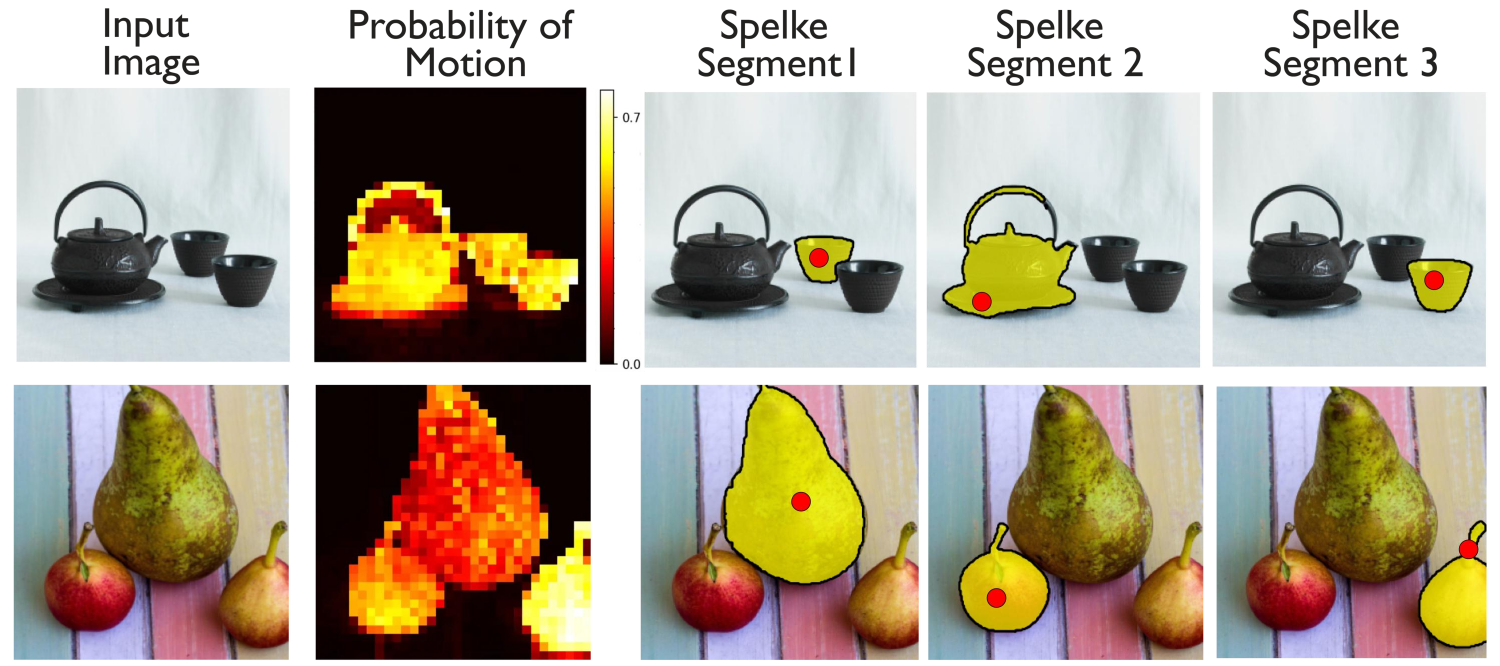}
  \caption{\textbf{Illustration of unprompted Spelke segment discovery using \lrasseg.} The corresponding discovered segments are highlighted, demonstrating the ability of \lrasseg to automatically identify every physically coherent, movable entity in the scene without manual prompts.}
  \label{fig:autoseg_suppl}
  \vspace{-1.0em}
\end{figure}

\textbf{Task.} So far, we have quantified how well our method discovers Spelke segments given point prompts. However, as we discussed in Section~\ref{sec:autoseg}, it is often desirable to automatically discover every Spelke segment in the scene. We show illustrative examples of discovered segments using the auto-discovery method proposed in Section~\ref{sec:autoseg} in Figure~\ref{fig:autoseg_suppl} and evaluate performance quantitatively in Table~\ref{tab:auto_seg}, on our \spelkeentity benchmark.

\textbf{Evaluation Metrics.} To evaluate segmentation quality, we compute Average Precision (AP), Average Recall (AR), F1-Score, and mIoU. Unlike point-prompted segmentation, where a point on a ground-truth object is provided and the model predicts a segment associated with that specific point, auto-segmentation outputs a set of segments without indicating which ground-truth objects they represent. To determine which predicted segment should be matched to which ground-truth segment, we compute the pairwise IoU matrix and apply the Hungarian method~\cite{kuhn1955hungarian} to find the best one-to-one matching, which is a necessary first step before we can compute our metrics. 

We then compute how many segments the model successfully detects by counting the number of predicted segments that are matched to ground-truth segments (i.e those that have an IoU greater than some threshold $\tau$). Given these detected segments, Average Precision (AP) measures the fraction of predicted segments that end up being matched and detected (i.e those that are in-fact Spelke objects). Average Recall (AR), by contrast, measures the fraction of ground-truth segments that are successfully detected by the model\footnote{Both AP and AR are averaged over multiple IoU thresholds in the range $\tau = (0.5, 0.99)$}. Intuitively, a model that predicts only a few high quality segments may achieve high precision but low recall as it may miss many segments, while a model that over-segments may boost recall at the cost of precision. The F1-Score balances these two metrics by computing their harmonic mean, providing an aggregate measure of segmentation performance. Finally, to assess how accurately the model predicts the boundaries of objects at the pixel-level, we use mIoU. For each GT segment \( g \in G \), where $G$ is the set of all GT segments, we use the IoU with its matched prediction (if any), or assign 0 if unmatched:
\[
\text{mIoU} = \frac{1}{|G|} \sum_{g \in G} \text{IoU}(g, \text{matched}(g))
\]

\textbf{Results.} Overall, we find that \lrasseg outperforms other self-supervised methods such as CutLER~\cite{wang2023cut} and ProMerge~\cite{li2024promerge} on most evaluation metrics. ProMerge slightly exceeds \lrasseg in AP due to its tendency to predict fewer segments than those in the GT---some of which align well with ground truth and thus boost precision---at the cost of lower recall, as some objects are missed.  For fairness, we report numbers only from the segment extraction stage for both CutLER and ProMerge, and not from their final distilled models.

\begin{table}[t]

  \centering
  \caption{\textbf{Quantitative evaluation of unprompted automatic segmentation across models on \spelkeentity.} We find that \lrasseg obtains competitive performance compared to existing self-supervised methods.}
  \label{tab:auto_seg}
  \small
  \setlength{\tabcolsep}{4pt}
  \begin{tabular}{lccccc}
    \toprule
      & SAM2~\cite{ravi2024sam2} & CutLER~\cite{wang2023cut} & ProMerge~\cite{li2024promerge} &  \lrasseg \\
    \midrule
    AP    & 0.11 & 0.41 & \textbf{0.42}  & 0.35   \\
    AR    & 0.62 & 0.32 & 0.34  & \textbf{0.46}   \\
    mIoU  & 0.68 & 0.42 & 0.43 & \textbf{0.57}   \\
    F1-score  & 0.17 & 0.34 & 0.36 & \textbf{0.38}  \\
    \bottomrule
  \end{tabular}
\end{table}

Compared to supervised methods like SAM2~\cite{ravi2024sam2}, \lrasseg achieves a higher F1 score, although its AR and mIoU are lower. This is largely expected: SAM often over-segments scenes based on texture and semantic cues, leading to multiple masks for a single Spelke object. While this increases the likelihood that at least one segment aligns with ground truth (improving recall), it reduces interpretability for physical reasoning, as these segments do not always capture true pixel co-movement.  This makes them less useful for downstream robotic applications which could benefit from such structure extractions. 

Interestingly, while our method outperforms SAM on point-prompted segmentation metrics, this advantage does not fully translate to the automatic setting. This suggests that when given the right point prompts, \lrasseg is capable of producing high-quality segments.  However, in their absence, performance may degrade due to occasional poor flow rollouts from sub-optimal point samples or to limitations in the current clustering strategy. This points towards a promising future direction: distilling our automatically discovered segments into a segmentation architecture like SAM could potentially factor out noise and improve performance, in the same spirit as distillation pipelines used in prior work~\cite{wang2023cut, hamilton2022unsupervisedsemanticsegmentationdistilling, li2024promerge}.

\subsection{Using Spelke segments for object manipulation}
\label{sec:obj_man}

\textbf{Task.} We consider the task of object-centric scene editing, where a user clicks a point on an object and provides an edit prompt specifying a 2D or 3D transformation. The object mask is generated from this point selection using a segmentation model. Here, we will present evidence that realistic edits require masks that reflect physically movable entities such as Spelke segments.

\textbf{Dataset.} To evaluate the utility of \lrasseg segments for object manipulation, we use 3DEditBench, recently introduced in~\cite{lee20253d}. The benchmark contains 100 real world images with associated point prompts, 3D transformation and resulting ground truth edited images with the transformation applied. It comprises of a diverse range of object types undergoing physical changes such as rotations, translations, and inter object occlusions.

\textbf{Baselines.} We evaluate our segments within several widely used image editing pipelines, including Lightning Drag~\cite{shi2024lightningdraglightningfastaccurate}, DiffusionHandles~\cite{pandey2024diffusionhandles}, and the recently introduced Diffusion-as-Shader model~\cite{gu2025diffusionshader3dawarevideo}, which demonstrated impressive performance on object manipulation tasks. We also evaluate the \lrasd model~\cite{lee20253d}, a state-of-the-art model for image editing, built on the \lras framework. For each method, we compare edits using SAM masks versus our \lrasseg  segments to isolate the effect of segmentation quality on edit realism and physical plausibility.

\begin{figure}[t!]  
  \centering
  \includegraphics[width=\linewidth]{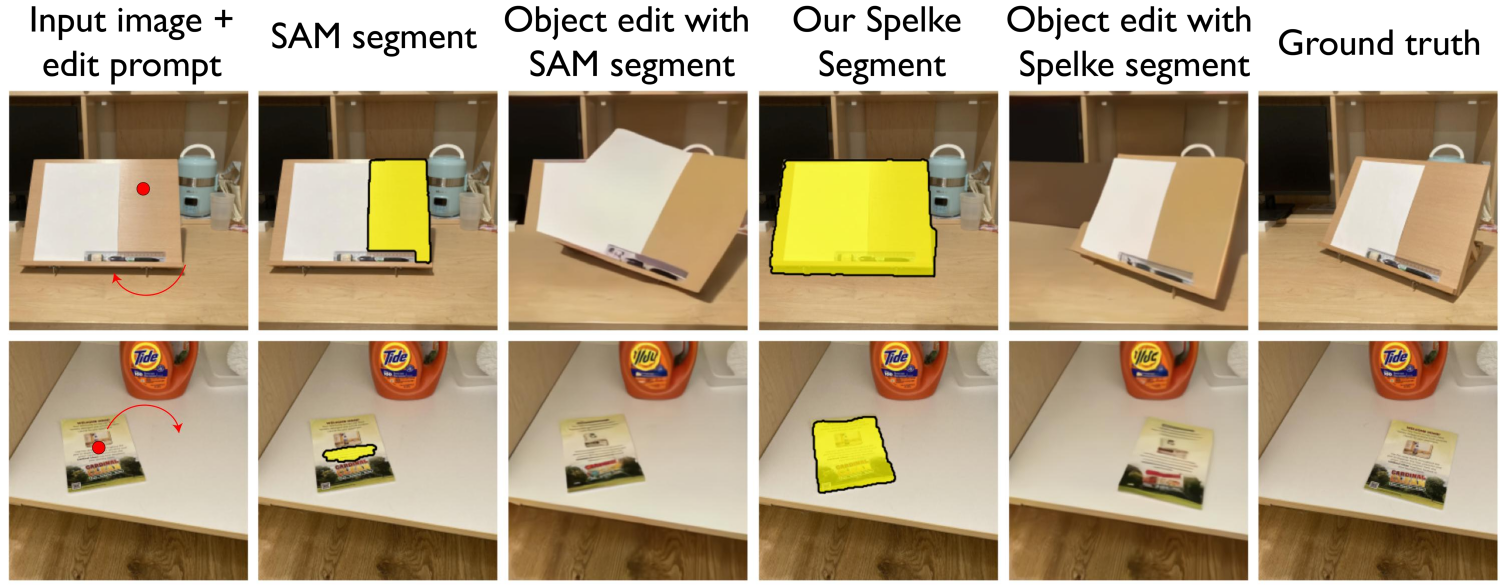}
  \vspace{-0.5em}
  \caption{\textbf{Qualitative comparisons of scene edits using SAM masks versus \lrasseg segments.} Each row shows the original image, the user click location, and the resulting edited image using segments from different methods. Note that while the results shown here are computed using the state-of-the-art \lrasd model for object editing, our segments are agnostic of the specific editing method used and can improve the results of any editing model as shown in Table~\ref{tab:obj_manipulation_accuracy}. Additional results are reported in the attached supplementary material.}
  \label{fig:object_manip_results}
  \vspace{-1.0em}
\end{figure}

\begin{table}[t!]
\centering
\caption{
\textbf{Quantitative evaluation of edit quality across segmentation methods and editing pipelines.} We report results for edits generated using SAM versus \lrasseg segments across four editing models. Lower $\downarrow$ is better, higher $\uparrow$ is better.
}
\label{tab:obj_manipulation_accuracy}
\begin{tabular}{llccccc}

\toprule
Method & Segment & MSE $\downarrow$ & PSNR $\uparrow$ & LPIPS $\downarrow$ & SSIM $\uparrow$ & EA $\uparrow$ \\
\midrule
\multirow{2}{*}{LRAS-3D~\cite{lee20253d}} 
  & \lrasseg & \textbf{0.009} & \textbf{21.64} & \textbf{0.213} & \textbf{0.698} & \textbf{0.776} \\
& SAM  & 0.013 & 20.17 & 0.255 & 0.685  & 0.633 \\
\midrule
 \multirow{2}{*}{LightningDrag~\cite{shi2024lightningdraglightningfastaccurate} }  & \lrasseg & \textbf{0.017 }& \textbf{19.16} & \textbf{0.195} & \textbf{0.672} & \textbf{0.679} \\
  & SAM  & 0.020 & 18.18 & 0.241 & 0.658 & 0.536 \\
\midrule
\multirow{2}{*}{DiffusionHandles~\cite{pandey2024diffusionhandles}} 
  & \lrasseg & \textbf{0.024} & \textbf{17.42} & \textbf{0.364} & \textbf{0.555} & \textbf{0.576} \\
  & SAM  & 0.031 & 16.15 & 0.419 & 0.526 & 0.495 \\
\midrule
\multirow{2}{*}{DiffusionAsShader~\cite{gu2025diffusionshader3dawarevideo}} 
  & \lrasseg & \textbf{0.015} & \textbf{19.29} & \textbf{0.194} & \textbf{0.707} & \textbf{0.640} \\
  & SAM  & 0.019 & 18.20 & 0.253 & 0.682 & 0.503  \\
\bottomrule
\end{tabular}
\end{table}

\textbf{Metrics.} While standard metrics like PSNR, SSIM, and LPIPS capture image quality, prior work~\cite{pandey2024diffusionhandles} has shown they often fail to reflect edit accuracy. To address this, they introduced the Edit Adherence (EA) metric, which measures how well the transformed object aligns with ground truth by computing the IoU between ground truth and predicted segments in the edited image. We report both this metric as well as standard image quality metrics. 

\textbf{Qualitative and Quantitative Comparisons.}
We find that \lrasseg segments consistently outperforms SAM, yielding physically grounded segments that improve realism across diverse image editing models (see Table~\ref{tab:obj_manipulation_accuracy}). In contrast, SAM-generated masks capture only sub-parts of objects, resulting in fragmented or implausible edits (see Figure~\ref{fig:object_manip_results}).

\begin{figure}[b!]
  \centering
  \includegraphics[width=\textwidth]{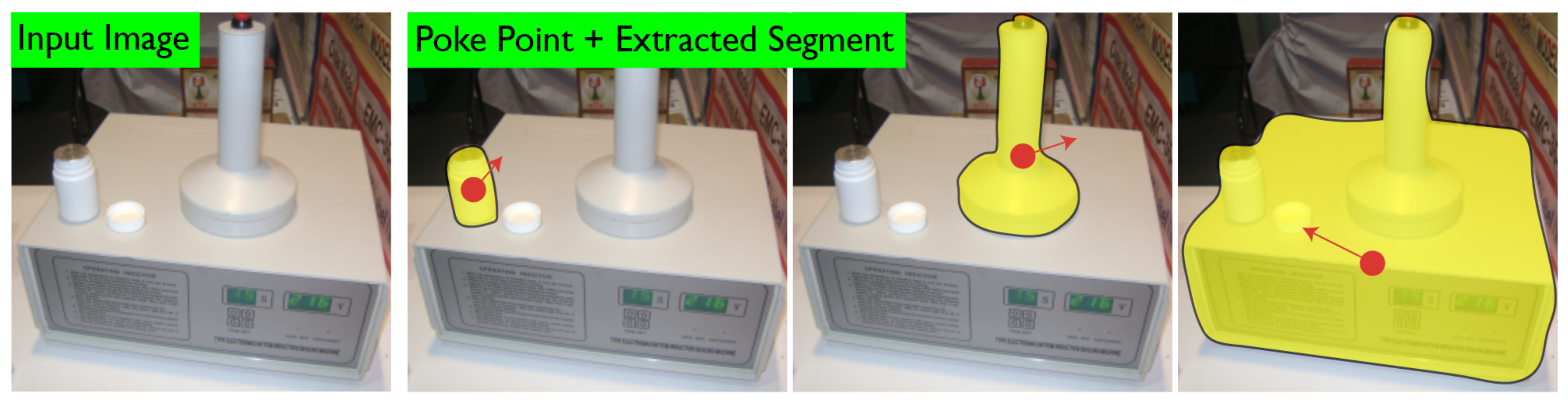}
  \vspace{-1em}
  \caption{\textbf{Support relationship understanding capabilities.} When applying a virtual poke to an object, the extracted Spelke segment includes both the directly contacted object and all the objects it physically supports, which implies an implicit understanding of the support hierarchy within a scene.} 
  \label{fig:support_relationships}
  \vspace{-1.0em}
\end{figure}

\section{Conclusion \& Future Work}
\label{sec:discussion}

In this paper, we show how a class of self-supervised visual world models---trained to predict plausible motion from input RGB images---can be used to discover motion-defined Spelke object entities from static images via zero-shot statistical counterfactual probing. To evaluate this approach, we introduce a new benchmark, \spelkeentity, which measures this capability, and find that our model, \lrasseg achieves superior results in comparison to both supervised and self-supervised segmentation methods on this benchmark.

While Spelke segments have largely been explored in cognitive science, we show in this paper that they align well with the kinds of abstractions needed for physically grounded computer vision and robotics tasks, such as selecting and manipulating coherent parts of a scene. When applied to the 3DEditBench object manipulation benchmark, Spelke segments enabled more physically plausible editing as they reflect what truly moves together in the scene. On the other hand, we found that models like SAM often split up or combine objects in ways that are inconsistent with how they move, resulting in segments that may be less useful when the goal is to physically manipulate objects.

\begin{figure}[t!]
  \centering
  \includegraphics[width=\textwidth]{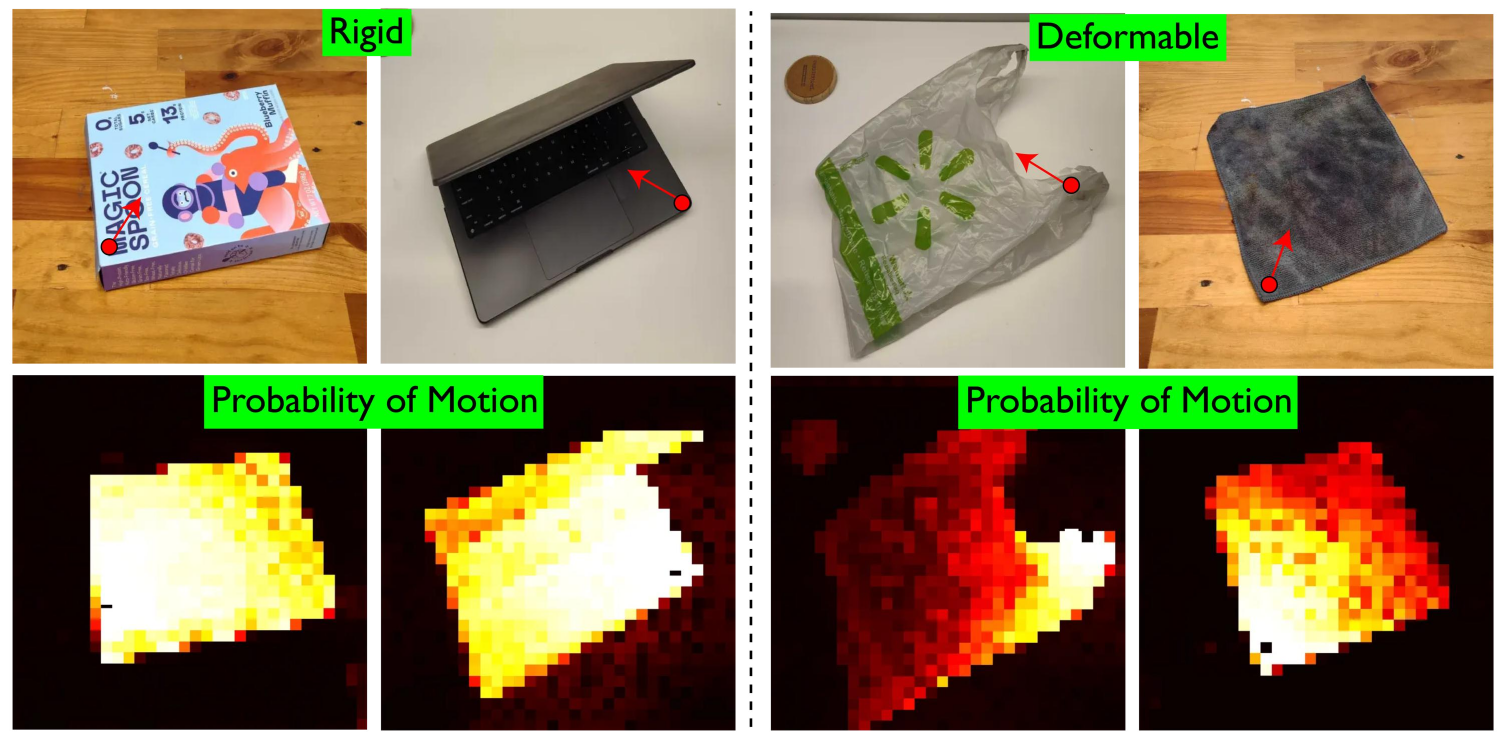}
  \vspace{-1em}
  \caption{\textbf{Emergent material property understanding capabilities.} We find that motion probability maps are uniform in case of rigid objects, but more localized near the virtual poke for deformable objects. This can potentially enable the discovery of material properties.}
  \label{fig:materials}
  \vspace{-1.0em}
\end{figure}

Looking ahead, we observe that causal probe-based structure extractions from \lrasseg may also offer a pathway to inferring other properties of the scene beyond segmentation. We find that segments from \lrasseg can reveal support relationships between objects. As shown in Figure~\ref{fig:support_relationships}, when virtually poking an object at the bottom of a stacked structure, the extracted segment includes every entity that the object physically supports. Additionally, as illustrated in Figure~\ref{fig:materials}, the $p_\text{motion}$ maps produced can be used to infer physical attributes such as rigidity or material type. For instance, rigid objects like laptops and cardboard boxes tend to exhibit a uniform probability across the segment, while deformable objects such as cloth and plastic covers often show more localized motion responses near the poke point. Exploring this connection between motion response patterns and physical properties is a promising direction for future work.

Though our focus in this paper has been on human-centric macroscopic physical scenes, the underlying philosophy of using predictive models to uncover causal and structural patterns through probing could open new avenues for data-driven discovery in other domains where humans have less direct intuition about the nature of objecthood. For example, in medical imaging, a model trained on time-lapse microscopy might help identify cohesive intra-cellular structures or track morphological changes, while in astrophysics, models trained on galaxy evolution data could be probed to discover gravitationally bound systems.

\clearpage

\section*{Acknowledgements}

This work was supported by the following awards: Simons Foundation grant SFI-AN-NC-GB-Culmination-00002986-05, National Science Foundation CAREER grant 1844724, National Science Foundation Grant NCS-FR 2123963, Office of Naval Research grant N00014-20-1-2589, ONR MURI N00014-21-1-2801, ONR MURI N00014-24-1-2748, and ONR MURI N00014-22-1-2740. We also thank Stanford HAI, Stanford Data Science, the Marlowe team, and the Google TPU Research Cloud team for providing computing support.

\vspace{1em}
{\centering\large\textbf{--- Supplementary Materials ---}\par}
\vspace{1em}


In this supplementary material, we provide additional qualitative results and further details about our architecture. This document is organized as follows:

\begin{itemize}
    \item \textbf{Section~\ref{sec:additionalqual}: Additional Qualitative results.}   
    \begin{itemize}
        \item More illustrations showing probability of motion maps (i.e $p_{\text{motion}}$) and expected direction of motion maps (i.e $\mathbb{E}_{\text{disp}}$) on real-world images and how they are used for Spelke segment discovery (Section~\ref{sec:addional_spelke_seg}). 
        \item Additional point-prompted segmentation results (Section~\ref{sec:addional_pointprompt}).
        \item More examples of downstream object manipulation using our Spelke segments (Section~\ref{sec:addional_obj_manip}).
    \end{itemize}

    \item \textbf{Section~\ref{sec:modelarch}: Further Architectural Details}  
\end{itemize}
 
\section{Additional Qualitative Results}
\label{sec:additionalqual}

\subsection{Additional Examples illustrating our Spelke Segment Discovery Algorithm}
\label{sec:addional_spelke_seg}
\begin{figure}[htbp]  
  \centering
  \includegraphics[width=0.9\textwidth]{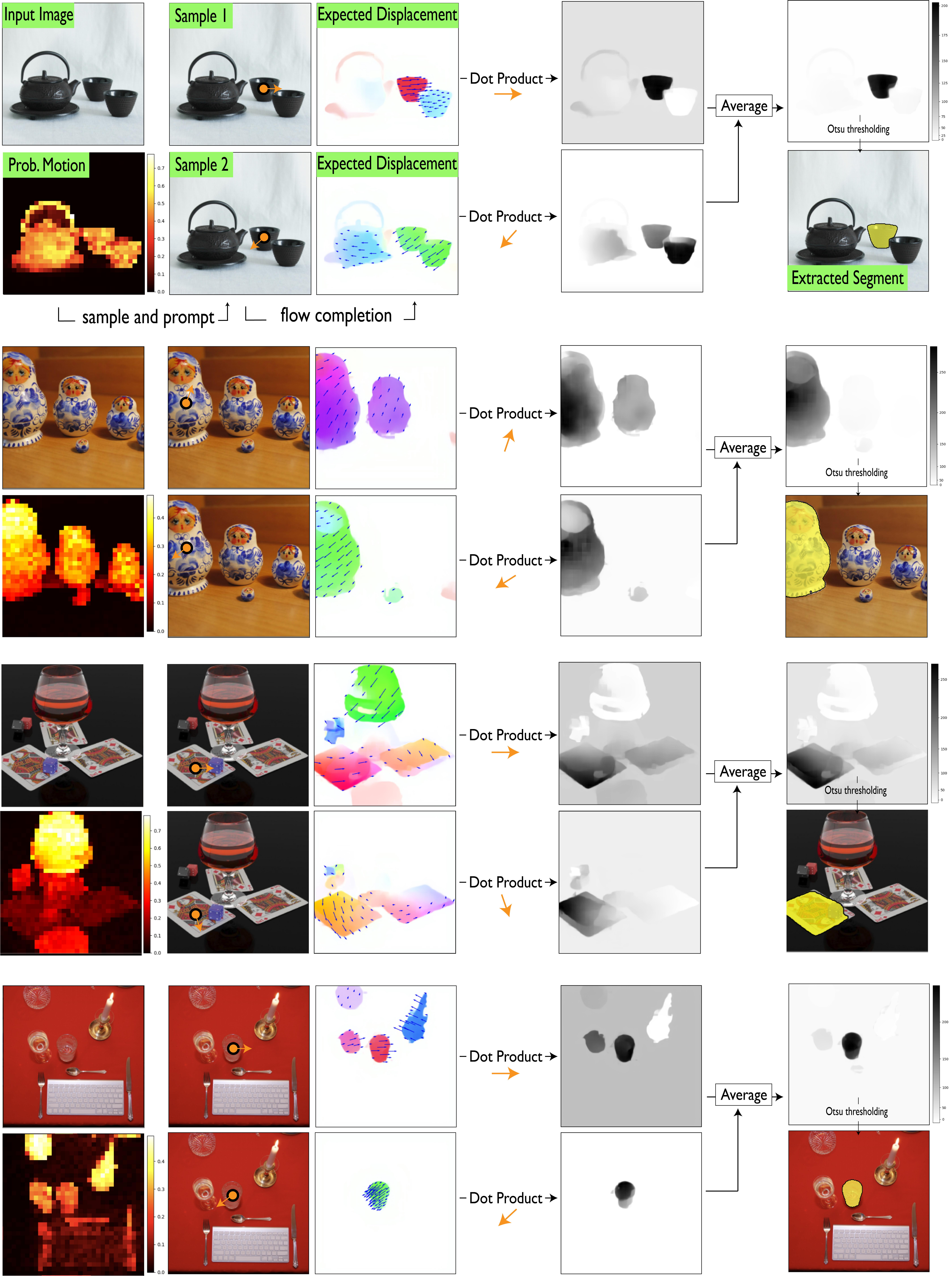}
  \vspace{-0.5em}
  \caption{\textbf{More illustrations of our Spelke segment discovery algorithm}. To discover movable objects, we apply multiple virtual pokes at locations sampled from the $p_{\text{motion}}$ map (column 2). While the model consistently propagates flow across the poked object (column 3), it also generates unprompted flow on other objects. However, since this unprompted flow varies across pokes and typically diverges in direction from the input poke, it gets suppressed when averaging the dot product (column 4) and helps us isolate independently movable entities as shown in the last column. Note that we average across 5 pokes, but only show two rows here for brevity.}
  \label{fig:point_seg_process}
  \vspace{-1.0em}
\end{figure}

In Section~\ref{sec:obj_discovery} and Figure~\ref{fig:spelke_discovery} of the main paper, we outlined our algorithm for discovering Spelke segments by simulating virtual pokes and aggregating directional flow responses. Here, in Figure~\ref{fig:point_seg_process} we provide additional examples visualizing the full process—from poke point sampling to segment extraction—highlighting the emergence of coherent, manipulable object masks. These visualizations further support the robustness and consistency of our flow-based grouping method across a diverse range of objects and scenes. These results also include more examples of probability of motion maps and expected direction of motion maps. However, unlike Figure~\ref{fig:spelke_discovery} of the main paper which only showed these maps generated in parallel mode, here we depict the expected direction of motion maps computed using multiple autoregressive rollouts (i.e. $\mathbb{E}_{\text{seq}}$)

\subsection{Point-Prompted Segmentation Results}
\label{sec:addional_pointprompt}
\begin{figure}[htbp]  
  \centering
  \includegraphics[width=0.9\textwidth]{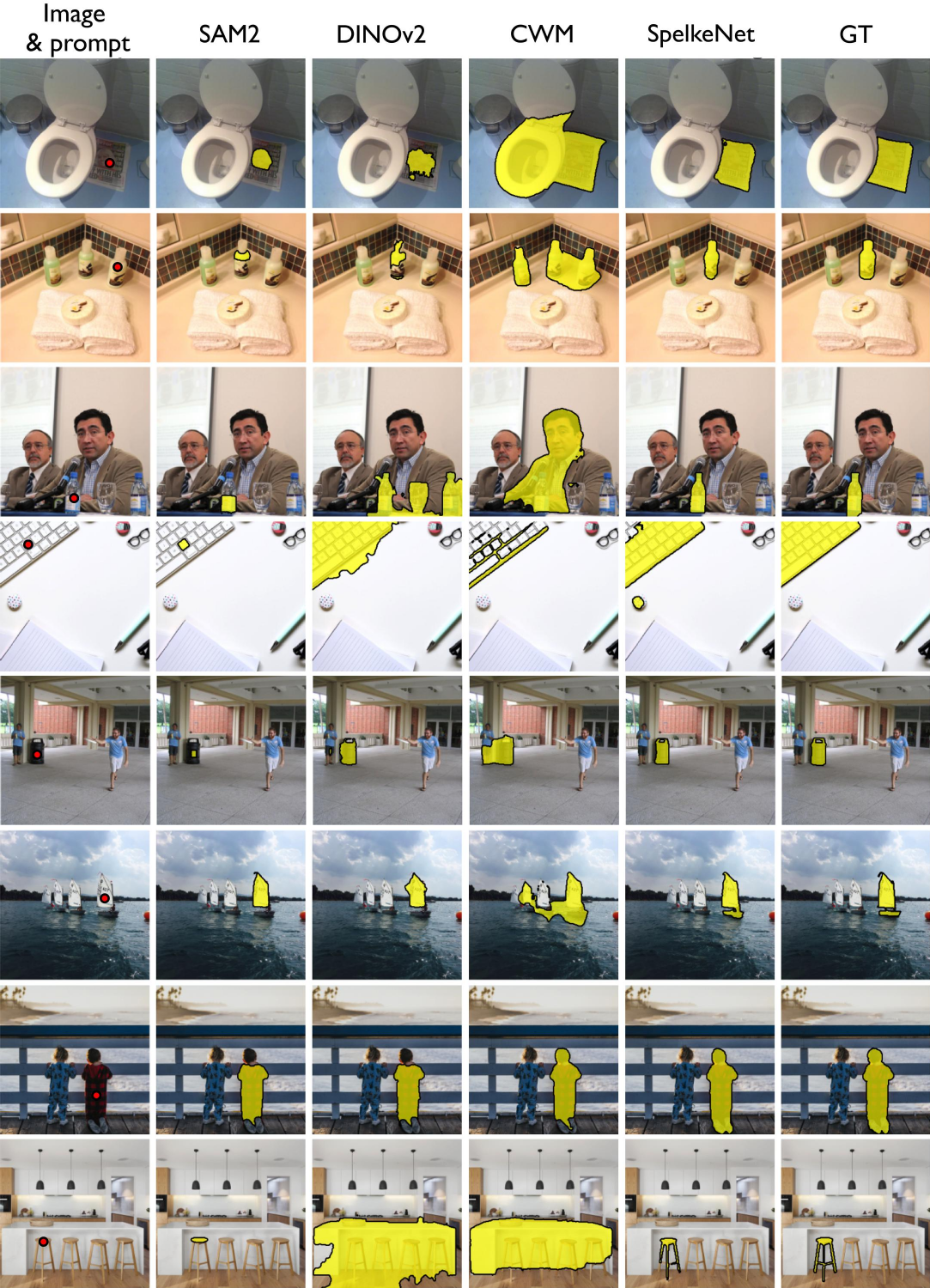}
  \vspace{-0.5em}
  \caption{\textbf{Additional qualitative results for point-promoted segmentation across models.} \lrasseg yields sharper, more ``Spelke-like'' segments compared to SAM2, DINO, and CWM.}
  \label{fig:point_seg_supplementary}
  \vspace{-1.0em}
\end{figure}

In Section~\ref{sec:point_prompted} of the main paper, we evaluated segmentation quality under point-prompted settings on our SpelkeEntitySeg benchmark \spelkeentity. Here, we present additional qualitative results comparing our method to baselines including SAM2~\cite{ravi2024sam2}, DINOv2~\cite{oquab2023dinov2}, and CWM~\cite{bear2023unifyingmachinevisioncounterfactual}. As shown in Figure~\ref{fig:point_seg_supplementary}, our method consistently produces cohesive and physically plausible segments, in contrast to alternatives that often fragment objects or include extraneous background.

\subsection{Object Manipulation Using Predicted Segments}
\label{sec:addional_obj_manip}
\begin{figure}[htbp]  
  \centering
  \includegraphics[width=0.9\textwidth]{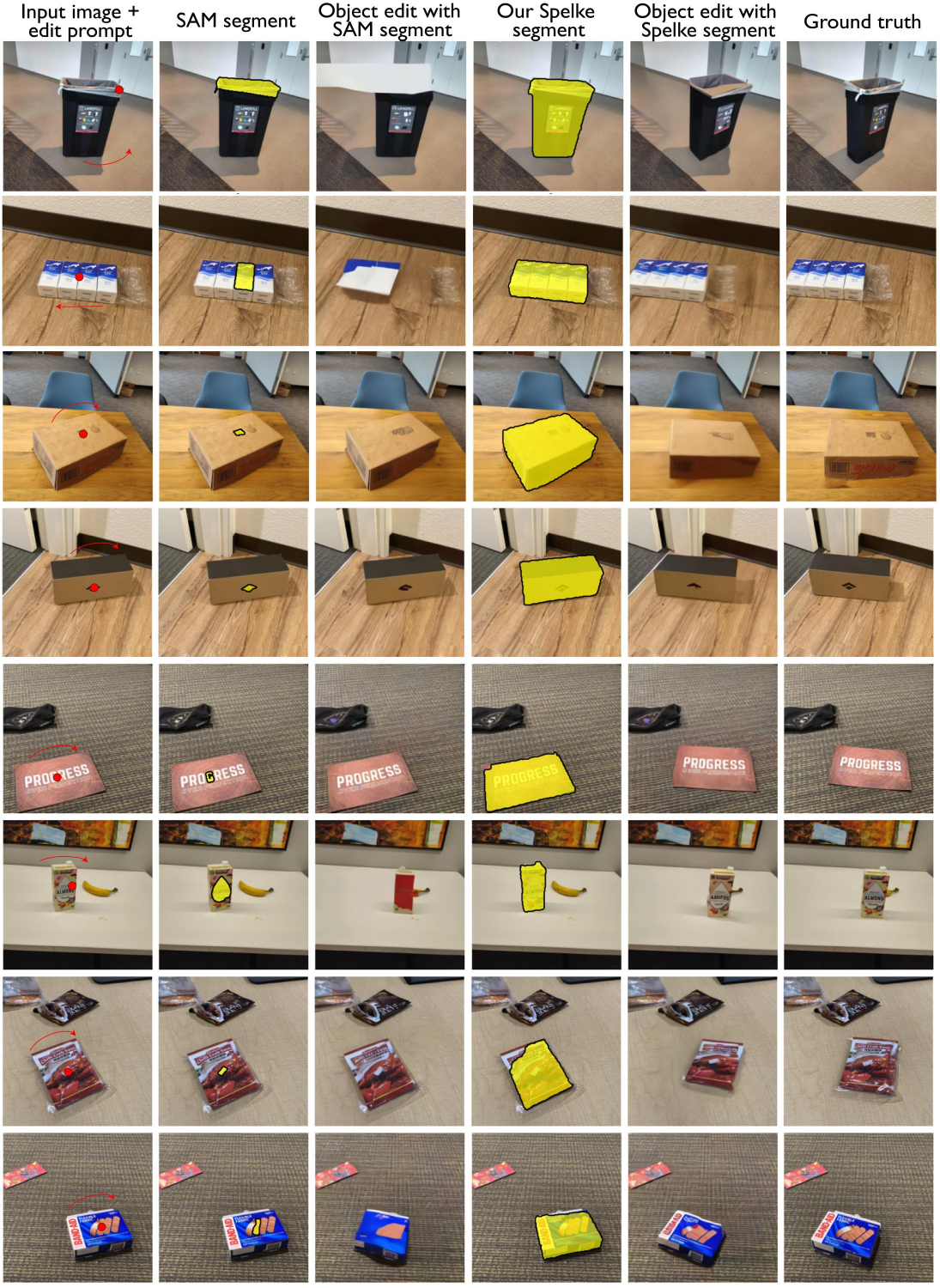}
  \vspace{-0.5em}
  \caption{\textbf{Additional qualitative comparisons of scene edits using SAM masks versus \lrasseg segments.} Each row shows the original image, the user click location, and the resulting edited image using different segmentation methods. }
  \label{fig:suppl_obj_manipulation}
  \vspace{-1.0em}
\end{figure}

We previously demonstrated in Section~\ref{sec:obj_man} of the main paper, the importance of physically grounded segmentation for object manipulation. Here, in Figure~\ref{fig:suppl_obj_manipulation} we include further qualitative comparisons of edits generated using our predicted segments versus those from SAM~\cite{ravi2024sam2}. As illustrated, segments aligned with Spelke objecthood significantly improve edit realism, spatial coherence, and transformation consistency across multiple 3DEditBench~\cite{lee20253d} examples.


\section{\lrasseg Specifications}
\label{sec:modelarch}

\subsection{Model Architecture}
LRAS~\cite{lee20253d} is a generative visual world model that predicts plausible optical flow fields conditioned on an RGB frame. It is a 7 billion-parameter autoregressive transformer (standard LLaMA architecture \cite{touvron2023llamaopenefficientfoundation}, 32 layers, 4096 embed dimensions, 32 attention heads) that operates over local patch tokens and can generate the predicted flow field in any order, sequentially or in parallel. 

\subsection{RGB and Flow Quantization}
The LRAS pipeline starts with a lightweight convolutional auto-encoder which quantizes each 4 × 4 pixel patch into an independent 16-bit code, yielding a 65,536-token vocabulary for RGB images (a second, similar quantizer is used to quantize optical-flow patches). 

\subsection{Enabling both locality and random access}
During serialization of the RGB and flow tokens into a 1D causal sequence, the model inserts special pointer tokens that tell the decoder which patch location to fill next, letting it generate images in an arbitrary or explicitly user-defined order instead of the usual raster scan. This locality plus random-access design promotes compositionality and gives every patch equal causal power. It also allows for fully parallel decoding, illustrating the current best estimate of the model's prediction at any step during the decoding (as described in Section~\ref{sec:seq_design} in the main paper).

\subsection{Dataset preparation}
The model is pre-trained on BVD (Big Video Dataset~\cite{lee20253d}) - a 7k hour dataset of diverse Internet videos mixed with standard 3-D vision datasets such as ScanNet++ \cite{yeshwanth2023scannetpp}, CO3D \cite{reizenstein2021co3d}, RealEstate-10K \cite{zhou2018realestate10k} and standard video datasets such as Kinetics \cite{kay2017kinetics}, SomethingSomethingv2 \cite{goyal2017something} and OpenX embodiment~\cite{open_x_embodiment_rt_x_2023}. Camera pose information is provided to the model whenever available in the dataset, and optical flow for every frame pair is computed with the SeaRAFT \cite{wang2024searaft} model and quantized.

\subsection{Training details}
The model is trained with sequence lengths of 4096 and a batch size of 512 for 200k steps with next-token cross-entropy loss. The quantizers themselves are first trained on Kinetics-400 frames with simple L2 reconstruction loss. The model was trained on 64 H100 GPUs for approximately 14 days.


\clearpage


\bibliographystyle{unsrtnat}
\bibliography{references}

\end{document}